\pdfoutput=1

\documentclass[11pt]{article}

\usepackage[final]{acl}
\usepackage{booktabs}
\usepackage{amsmath}
\usepackage{floatrow}
\usepackage{float}

\usepackage{times}
\usepackage{latexsym}
\usepackage{amssymb}

\usepackage[T1]{fontenc}
\usepackage{longtable}

\usepackage[utf8]{inputenc}

\usepackage{microtype}

\usepackage{inconsolata}

\usepackage{graphicx}

\usepackage{cleveref}
\PassOptionsToPackage{hyphens}{url}\usepackage{hyperref}       
\usepackage{nicefrac}       
\usepackage{microtype}      
\usepackage{cleveref}
\usepackage{wrapfig}

\usepackage{multirow}
\usepackage{subcaption}

\usepackage{fontawesome}

\newcommand{\customurl}[1]{}

\title{Geometric Signatures of Compositionality \\Across a Language Model's Lifetime}

\author{
 \textbf{Jin Hwa Lee\textsuperscript{*1}},
 \textbf{Thomas Jiralerspong\textsuperscript{*2,3}},
 \textbf{Lei Yu\textsuperscript{4}},
 \textbf{Yoshua Bengio\textsuperscript{2,3}},
 \textbf{Emily Cheng\textsuperscript{5}}
\\
\\
 \textsuperscript{1}University College London,
 \textsuperscript{2}Université de Montréal,
 \textsuperscript{3}Mila,
 \textsuperscript{4}University of Toronto,
 \textsuperscript{5}Universitat Pompeu Fabra
\\
 \small{
   \textbf{*Equal contribution, co-first authors.}}
   \\
   \small{\textbf{
   Correspondence:} \href{mailto:jin.lee.22@ucl.ac.uk}{jin.lee.22@ucl.ac.uk} and \href{mailto:thomas.jiralerspong@mila.quebec}{thomas.jiralerspong@mila.quebec} 
   and \href{mailto:emilyshana.cheng@upf.edu}{emilyshana.cheng@upf.edu}
 }
}

\begin{document}
\maketitle
\begin{abstract}
By virtue of linguistic compositionality, few syntactic rules and a finite lexicon can generate an unbounded number of sentences. That is, language, though seemingly high-dimensional, can be explained using relatively few degrees of freedom. An open question is whether contemporary language models (LMs) reflect the intrinsic simplicity of language that is enabled by compositionality. We take a geometric view of this problem by relating the degree of compositionality in a dataset to the intrinsic dimension ($I_d$) of its representations under an LM, a measure of feature complexity. We find not only that the degree of dataset compositionality is reflected in representations' $I_d$, but that the relationship between compositionality and geometric complexity arises due to learned linguistic features over training. Finally, our analyses reveal a striking contrast between nonlinear $I_d$ and linear dimensionality, showing they respectively encode semantic and superficial aspects of linguistic composition. 

\noindent \faicon{github} {\href{https://github.com/jinhl9/llm-compositionality-lifetime}{{\fontfamily{cmss}\selectfont jinhl9/llm-compositionality-lifetime
}}}

\end{abstract}

\section{Introduction}

\label{sec:intro}

Compositionality, the notion that the meaning of an expression is constructed from that of its parts and syntactic rules \citep{Szabó_2024}, enables linguistic atoms to locally combine to create global meaning \citep{Frege1948-FREUSU-3,Chomsky1999DerivationBP}. Then, a rich array of meanings at the level of a phrase may be explained by simple rules of composition. If a language model is a good model of language, we expect its internal representations to reflect the latter's relative simplicity. That is, representations should reflect the \emph{manifold hypothesis}, or that real-life, high-dimensional data lie on a low-dimensional manifold \citep{Goodfellow-et-al-2016}. 

The dimension of this manifold, or \emph{nonlinear intrinsic dimension} ($I_d$), is the minimal number of degrees of freedom needed to describe it without information loss \citep{Goodfellow-et-al-2016,Campadelli_Casiraghi_Ceruti_Rozza_2015}. The manifold hypothesis is attested for linguistic representations: LMs are found to compress inputs to an $I_d$ orders-of-magnitude lower than their ambient dimension  \citep{cai2021isotropy,Cheng:etal:2023,valeriani2023the}. 

A natural question is whether the inherent simplicity of linguistic utterances, enabled by compositionality, manifests in representation manifolds of low complexity, described by the manifolds' $I_d$. For example, consider the set of possible phrases generated by the syntactic structure \texttt{[DET][N][V]}. While the possible utterances have combinatorial complexity, they are generated by only three degrees of freedom. An LM that can handle this syntactic structure should be able to both \textcolor{MidnightBlue}{(1)} represent all combinations; \textcolor{BrickRed}{(2)} encode the few degrees of freedom underlying them, or their \emph{degree of compositionality}. We ask, which geometric aspects of the representation space encode \textcolor{MidnightBlue}{(1)} and \textcolor{BrickRed}{(2)}? 

The manifold hypothesis predicts the $I_d$ of LM representations to encode \textcolor{BrickRed}{(2)} \citep{Recanatesi_Farrell_Lajoie_Deneve_Rigotti_Shea-Brown_2021}. But, so far in the literature, while some linguistic categories have been found to occupy low-dimensional linear subspaces \citep{hernandez-andreas-2021-low,Mamou_Le_Rio_Stephenson_Tang_Kim_Chung_2020}, an explicit link between degree of compositionality and representational dimensionality has not been established. To bridge this gap, in large-scale experiments on causal LMs and a custom dataset with tunable compositionality, we provide first insights into the relationship between the degree of compositionality of inputs and geometric complexity of their representations over the course of training. 

\paragraph{Outline} On both our controlled datasets and naturalistic stimuli, we reproduce the finding \citep{cai2021isotropy} that LMs represent linguistic inputs on low-dimensional nonlinear manifolds. We also show for the first time that LMs expand representations into high-dimensional \emph{linear} subspaces, concretely, that \textbf{(I)} nonlinear $I_d$ and linear dimensionality scale differently with model size. We show that feature geometry is relevant to behavior over training, in that \textbf{(II)} LMs' feature complexity tracks a phase transition in linguistic competence. Different from past work, we consider two different measures of dimensionality, nonlinear and linear, showing that they encode complementary information about compositionality. Our key result is that \emph{representations' geometric complexity reflects dataset complexity}. However, \emph{nonlinear} $I_d$ encodes \textcolor{BrickRed}{(2) meaningful compositionality} while \emph{linear} dimensionality encodes \textcolor{MidnightBlue}{(1) superficial input complexity}, in a way that arises over training: \textbf{(III)} nonlinear $I_d$ reflects superficial complexity as an inductive bias of LM's architecture, but meaningful compositional complexity by the end of training. Instead, \textbf{(IV)} linear dimensionality, not $I_d$, encodes superficial complexity throughout training. Overall, results show a contrast between linear and nonlinear feature complexity that suggests their relevance to \textcolor{MidnightBlue}{form} and \textcolor{BrickRed}{meaning} in how LMs process language.

\section{Background}
\paragraph{Compositionality} Compositionality is the notion that \emph{the meaning of a complex expression can be constructed from that of its parts and how they are structured} \citep{Frege1948-FREUSU-3,Partee_1995}. While compositionality, per this definition, is a general property of \emph{language}, it spans various interpretations that we review here. In linguistic analysis, compositionality is typically measured at the level of a \emph{single sample}: for instance, ``silver table", whose meaning can be constructed intersectively from ``silver" and ``table", is more compositional than the idiom ``silver spoon" \citep{Labov19806PI}. Beyond single instances, compositionality has been characterized at the \emph{system level}, for instance, to measure the extent to which English realizes all possible word combinations \citep{Sathe_Fedorenko_Zaslavsky_2024} or to compare the structures of different communication systems \citep{Brighton_Kirby_2006,chaabouni-etal-2020-compositionality,elmoznino2024complexitybasedtheorycompositionality}. Beyond language itself, researchers have considered what it means for \emph{users or models of language} to be compositional \citep{Hupkes2019CompositionalityDH} by evaluating compositionality in terms of model behavior \citep{lake2018generalization,sysgen2019} and representations \citep{smolensky1990tensor,andreas2018measuring}.

We are interested in a system-level notion of compositionality that is quantified on a linguistic dataset, further detailed in \Cref{subsec:compositionality}. Then, rather than explicitly describe the LM's composition function~\citep{smolensky1990tensor}, we ask simply whether LMs preserve \emph{inputs'} degree of compositionality in their representations' geometric complexity.

\paragraph{Manifold hypothesis and low-dimensional geometry}
Deep learning problems appear high-dimensional, but research suggests that they have low-dimensional intrinsic structure. In computer vision, image data and common learning objectives live on low-dimensional manifolds \citep{li2018measuring,valeriani2023the,Psenka_Pai_Raman_Sastry_Ma_2024,Ansuini_Laio_Macke_Zoccolan_2019}. Similarly, LMs' learning dynamics occur in low-dimensional parameter subspaces \citep{aghajanyan-etal-2021-intrinsic,zhang-etal-2023-fine}. The parsimony that governs these models' latent spaces reduces learning complexity \citep{Cheng:etal:2023,pope2021the} and may arise from the training objective, seen in classification \citep{chung_2018} and sequential prediction \citep{Recanatesi_Farrell_Lajoie_Deneve_Rigotti_Shea-Brown_2021}. 

For language, the geometry of representations has been explored in several contexts. Prior work describes how concepts organize in representation space \citep{engels2024languagemodelfeatureslinear,park2025the,balestriero2024characterizing,Doimo_Serra_Ansuini_Cazzaniga_2024}; representational geometry can encode tree-like syntactic structures \citep{andreas2018measuring,murtycharacterizing,alleman-etal-2021-syntactic,hewitt-manning-2019-structural}; and certain linguistic categories like part-of-speech lie in low-dimensional linear subspaces \citep{Mamou_Le_Rio_Stephenson_Tang_Kim_Chung_2020,hernandez-andreas-2021-low}. Most similar to our work, \citet{Cheng:etal:2023} report the $I_d$ of representations over layers as a measure of feature complexity for natural language datasets, finding an empirical link between information-theoretic and geometric compression. But, our work is the first to explicitly relate the compositionality of inputs, a critical feature of language, to the degrees of freedom, or $I_d$, of their representation manifold. 

\paragraph{Language model training dynamics} Most research on LMs focuses on their final configuration at the end of training. Yet, recent work shows that learning dynamics provide cues as to how behavior arises~\citep{chen2024sudden,singh2024transient,tigges2024llm}. It has been found that, over training, LM weights become higher-rank \citep{abbe2023transformers} and their gradients more diffuse \citep{weber-etal-2024-interpretability}. 
Phase transitions are attested for some, but not all, aspects of language learning in LMs. Negative evidence includes that circuits involved in linguistic tasks are stable \citep{tigges2024llm} and gradually reinforced \citep{weber-etal-2024-interpretability} over training. Positive evidence includes that BERT's feature complexity tracks sudden syntax acquisition and drops in training loss \citep{chen2024sudden}, with similar findings for Transformers trained on formal languages \citep{lubana2024percolation}. Our work complements these results, exploring how the relationship between compositionality of inputs and complexity of their representations arises over training.

\section{Setup}

\subsection{Compositionality of a system} 
\label{subsec:compositionality}
We focus on a \emph{system-level} view of compositionality \citep{Sathe_Fedorenko_Zaslavsky_2024}: the extent to which a language system expresses all possible compositional meanings given a vocabulary. Consider, for example, systems \textbf{A} and \textbf{B} with the same syntactic structure \texttt{[DET][ADJ][N][V]} and vocabulary items:\vspace{1em} 
\scalebox{0.9}{
\begin{tabular}{ll}
        \textbf{A} & \textbf{B} \\
        \texttt{the blue dog runs} &  \texttt{the blue dog runs} \\
        \texttt{the red cat runs} & \texttt{the red cat talks}\\
        \texttt{the red dog talks} & \texttt{the blue dog talks} \\
        \texttt{the blue cat talks} & \texttt{the red cat runs}
\end{tabular}} \vspace{1.5ex}

In \textbf{A}, all adjectives, nouns, and verbs can co-occur, while in \textbf{B}, some adjective-noun pairs (``blue cat" and ``red dog") can never co-occur. Then, since more meanings can be created from the same vocabularies, we consider system \textbf{A} more compositional than system \textbf{B}. 

We see that a system's degree of compositionality is intimately linked to its \emph{degrees of freedom}. In the more compositional system \textbf{A}, there are three degrees of freedom: \texttt{[ADJ]}, \texttt{[N]} and \texttt{[V]}. In \textbf{B}, there are two: \underline{\texttt{[ADJ][N]}} and \texttt{[V]}. If a language model adheres to the manifold hypothesis, then these degrees of freedom in the inputs should be reflected in their representation manifold. Then, we expect ``system \textbf{A} is more compositional than \textbf{B}" to imply ``under an LM, \textbf{A}'s representations are more complex than \textbf{B}'s". Testing this hypothesis requires controlled datasets with different degrees of compositionality, detailed in the next section, and a measure of feature complexity, see \Cref{subsec:complexity-estimation}.

\subsection{Datasets} 
\label{subsec:datasets}
Following the intuition outlined in the previous section, we design several datasets who differ solely in their degree of compositionality. We later replicate key findings on The Pile \citep{gao2020pile}, a dataset of naturalistic language.

\subsubsection{Controlled grammar}
Our datasets contain sentences generated from a synthetic grammar. To create it, we set $12$ semantic categories and uniformly sample a $50$-word vocabulary for each category; categories' vocabularies are disjoint. We fix the syntactic structure by ordering the word categories: \\

\noindent

The \texttt{\textcolor{olive}{[quality$_1$.ADJ] [nationality$_1$.ADJ] [job$_1$.N]} \textcolor{Periwinkle}{[action$_1$.V]}} the \texttt{\textcolor{olive}{[size$_1$.ADJ] [texture.ADJ] [color.ADJ] [animal.N]}} then \texttt{\textcolor{Periwinkle}{[action$_2$.V]}} the \texttt{\textcolor{olive}{[size$_2$.ADJ] [quality$_2$.ADJ] [nationality$_2$.ADJ] [job$_2$.N]}.} \\

This produces sentences that are $17$ words long.
The order is chosen so that generated sentences are grammatical and that the adjective order complies with the accepted ordering for English \citep{dixon1976iwhere}. Vocabularies are chosen so that sentences are semantically coherent: for instance, if the agent is a person and patient is an animal, then verbs are constrained to permit, e.g., ``\textcolor{Periwinkle}{pets}", but not ``\textcolor{Periwinkle}{types}". To explore the effect of sequence length on feature complexity, we design four additional, similar grammars of lengths $\in \{5, 8, 11, 15\}$ words. Details for the sampling procedure, vocabularies, and syntactic structures are found in \Cref{app:grammar}.

\paragraph{Controlling compositionality} 
We modify the grammar to vary the dataset's degree of compositionality. To do so, we couple the values of $k$ contiguous word positions to produce a different dataset for each $k\in \{1\cdots 4\}$. That is, during data generation, $k$-grams are independently sampled, which constrains the degrees of freedom in each sentence to $l / k$ where $l$ is the number of categories. For instance, for the longest grammar, there are $l$=$12$ categories; in the $1$-coupled setting, each category is sampled independently, hence $12$ degrees of freedom; in the $2$-coupled setting, each bigram is sampled independently, hence $6$ degrees of freedom. Crucially, varying $k$ maintains the dataset's unigram distribution, only changing the dataset's $k$-gram distributions, or compositional complexity. 

Since the sentences in our datasets are \emph{literal} (form-meaning mappings are one-to-one), the trends we find when varying $k$ could be attributed to superficial complexity differences in wordforms, rather than in meanings. To ablate this effect, for each $k$, we create a parallel dataset of ``meaningless" inputs by randomly shuffling the words in each sequence. Shuffling preserves shallow distributional properties like length, sentence-level co-occurrences, and unigram frequencies (\Cref{fig:prompt} right). Notably, shuffled datasets are semantically vacuous, but their superficial complexities are still ordered in $k$; trends observed on shuffled datasets can only be due to changes in superficial complexity and not meaningful compositional complexity. 

If representations encode inputs' degree of compositionality, set by $k$ and whether sequences are shuffled, then we expect: (1) smaller $k$, meaning more degrees of freedom in the dataset, implies \emph{higher} feature complexity; (2) shuffling, which breaks semantic composition, entails \emph{lower} feature complexity. For each case in $k \in \{1 \cdots 4\} \times \{$coherent, shuffled$\}$, we randomly sample $5$ data splits of $10000$ sequences, reporting over splits. 

\paragraph{Measuring compositionality}  
As linguistic structure permits compression, it is common to measure the complexity of linguistic data using \emph{information-theoretic} quantities \citep{levy2008expectation,mollica2019humans,elmoznino2024complexitybasedtheorycompositionality}. Along these lines, we quantify compositionality of a dataset by its Kolmogorov complexity (KC). In algorithmic information theory, KC is the length of the shortest program in bits needed to generate the dataset. Datasets that are more compositionally complex (lower $k$, more degrees of freedom) are less compressible, thus have higher KC. While the true KC is intractable \citep{Cover2006}, we approximate it as others have~\citep{jiang-etal-2023-low}, using the lossless compression algorithm \texttt{gzip}. We estimate KC for $k\in \{1\cdots 4\} \times \{\text{coherent, shuffled}\}$ by the \texttt{gzip}-ped dataset size in kilobytes, then correlate it to feature complexity (\Cref{subsec:complexity-estimation}) for each layer. 

The connection between a dataset's KC and its degree of compositionality requires careful interpretation, as \texttt{gzip} is not semantics-aware: it only measures the compressibility of the superficial wordforms that make up the sequences. In coherent datasets, sequences are grammatical and literal; then, the superficial complexity measure returned by \texttt{gzip} directly proxies the degree of compositionality. For shuffled datasets, whose sequences are semantically vacuous, KC instead only measures wordforms' superficial combinatorial complexity.

\subsubsection{The Pile}
We focus on the synthetic grammar to tune compositionality in a surgical way. But, to ensure results are not an artifact of our prompts, we replicate experiments on The Pile, uniformly sampling $5$ random splits of $N$=$10000$ sequences that are $17$ words, the same length as sequences in the longest controlled grammar. We report results over splits. 

\subsection{Models}
\label{subsec:models}
We consider Transformer-based causal LMs Llama-3-8B (hereon Llama) \citep{llama3}, Mistral-v0.1-7B (Mistral) \citep{jiang2023mistral}, and Pythia models of sizes $\in \{$14m, 70m, 160m, 410m, 1.4b, 6.9b, 12b$\}$ \citep{biderman2023pythia}. Pythia is trained on The Pile, a natural language corpus spanning encyclopedic text, books, social media, code, and reviews \citep{gao2020pile}; Llama and Mistral's training data are not public, but likely similar. Models are trained to predict the next token given context, subject to a negative log-likelihood loss.

\paragraph{Pre-training analysis}
Pythia's intermediate training checkpoints are public \citep{biderman2023pythia}. For three intermediate sizes 410m, 1.4b, and 6.9b, we report Pythia's performance throughout pre-training on the set of evaluation suites provided by \citep{biderman2023pythia, eval-harness}, detailed in \Cref{app:tasks}. This encompasses a range of higher-level linguistic and reasoning tasks, from long-range text comprehension \citep{paperno-etal-2016-lambada} to commonsense reasoning \citep{bisk2020piqa}. The evolution of task performance provides a cue for the type of linguistic knowledge learned by the LM by a certain training checkpoint.

  \begin{figure*}
\floatbox[{\capbeside\thisfloatsetup{capbesideposition={right,top},capbesidewidth=0.34\textwidth}}]{figure}[\FBwidth]
{\caption{\textbf{Dimensionality over model size.} Mean layerwise $I_d$ (left) and $d$ (right) are shown for increasing hidden dimension. $I_d$ does not depend on hidden dimension $D$ (flat lines), but PCA $d$ scales linearly in $D$. Curves are averaged over 5 data splits $\pm$ 1 SD.}\label{fig:id_size_scaling}}
{\includegraphics[width=0.61\textwidth]{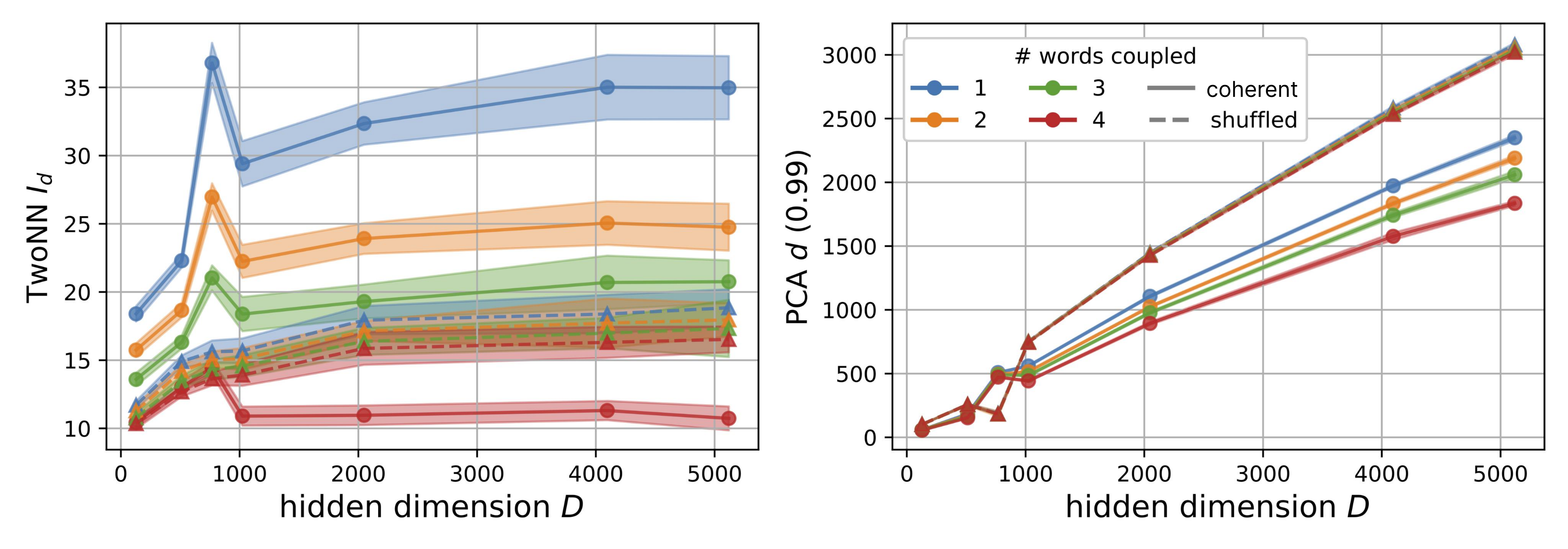}}
\end{figure*}

\subsection{Measuring feature complexity}
\label{subsec:complexity-estimation}
We are interested in how feature complexity reflects an input dataset's degree of compositionality. We consider representations in the Transformer's \emph{residual stream}~\citep{elhage2021mathematical}. As sequence lengths can slightly vary due to tokenization, as in prior work \citep{Cheng:etal:2023,Doimo_Serra_Ansuini_Cazzaniga_2024}, we aggregate over the sequence by taking the last token representation, as, due to causal attention, it is the only one to attend to the entire context.

For each layer and dataset, we measure feature complexity using both a nonlinear and a linear measure of dimensionality. Nonlinear and linear measures have key differences. The nonlinear $I_d$ is the number of degrees of freedom, or latent features, needed to describe the underlying manifold \citep{Campadelli_Casiraghi_Ceruti_Rozza_2015} (see \Cref{app:id}). This differs from the \emph{linear} effective dimension $d$, the dimension of the minimal linear subspace that explains representations' variance. We will use \emph{dimensionality} to refer to both nonlinear and linear estimates, specifying, when needed, $I_d$ as the nonlinear, $d$ as the linear, and $D$ as the LM's hidden dimension. 

\paragraph{Intrinsic dimension} We report the nonlinear $I_d$ using the TwoNN estimator \citep{Facco_d'Errico_Rodriguez_Laio_2017}. We choose TwoNN for several reasons: (1) it is highly correlated to other state-of-the-art estimators
\citep{Cheng:etal:2023,Facco_d'Errico_Rodriguez_Laio_2017}; (2) it relies on minimal assumptions of local uniformity up to the second nearest neighbor of a point, in contrast to other methods that impose stricter assumptions like global uniformity \citep{Albergante_Bac_Zinovyev_2019}; (3) TwoNN and correlated estimators are often used in the manifold estimation literature \citep{Cheng:etal:2023,pope2021the,chen2024sudden,tulchinskii2023intrinsic,Ansuini_Laio_Macke_Zoccolan_2019}. We focus on TwoNN in the main text, but test another estimator, \citealp{Levina_Bickel_2004}'s MLE, in \Cref{app:id_estimators}, confirming it is highly correlated.

TwoNN works as follows. Points on the underlying manifold are assumed to follow a locally homogeneous Poisson point process. Here, local refers to the neighborhood about each point $x$, up to $x$'s second nearest neighbor. Let $r_{k}^{(i)}$ be the Euclidean distance between $x_i$ and its $k$th nearest neighbor. Then, under the stated assumptions, distance ratios $\mu_i \triangleq r_2^{(i)} / r_1^{(i)} \in [1, \infty)$ follow the cumulative distribution function $F(\mu) = (1 - \mu^{-I_d})\mathbf 1 [\mu \geq 1]$. This yields the estimator $I_d = - \log (1 - F(\mu)) / \log \mu$. Lastly, given representations $\{x_i^{(j)}\}_{i=1}^N$ for LM layer $j$, $I_d^{(j)}$ is fit via maximum likelihood estimation over all datapoints.

\paragraph{Linear effective dimension} To estimate the linear effective dimension $d$, we use Principal Component Analysis (PCA) \citep{pca} with a variance cutoff of $99\%$. We focus on PCA in the main text; for other linear methods see \Cref{app:id_estimators}.


\section{Results}
We find representational dimensionality to systematically reflect compositionality over pre-training and model scale. 
We show that LMs represent linguistic data on low-dimensional, nonlinear manifolds, but in high-dimensional linear subspaces that scale with the hidden dimension. Then, we show that, over training, feature complexity tracks an LM's linguistic competence, where both exhibit a phase transition marking the emergence of syntactic and semantic abilities. Finally, we show that feature complexity reflects degree of compositionality, but that nonlinear $I_d$ encodes meaningful compositional complexity while linear $d$ encodes superficial complexity. We focus on Pythia 410m, 1.4b, and 6.9b here, with full results in the Appendix.



\subsection{Nonlinear and linear feature complexity scale differently with model size}
\label{sec:results-model-size}
We find, in line with \citet{cai2021isotropy,Cheng:etal:2023}, that LMs represent data on nonlinear manifolds of $I_d$ orders-of-magnitude lower than the embedding dimension. For the controlled dataset (\Cref{fig:id_size_scaling}) and The Pile (\Cref{fig:pile-id-model-size-app}), $I_d$ is $O(10)$ while linear $d$ and $D$ are $O(10^3)$. 

Our novel finding is that $d$ scales linearly in LM hidden dimension $D$, while $I_d$ is robust to it. We fit linear regressions from $\langle d\rangle_{\text{layer}}$ and $\langle I_d \rangle_{\text{layer}}$ to $D$ for each setting in $k\in \{1\cdots 4\} \times \{\text{coherent, shuffled}\}$. In all cases, $d \propto D$, see \Cref{fig:id_size_scaling} (right); all show a highly significant linear fit with $R>0.99$ and $p<0.005$ (\Cref{tab:linear_fitting_D_PCA_twonn_prompts}). Instead, $I_d$ stabilizes to a low range $O(10)$ regardless of $D$, see \Cref{fig:id_size_scaling} (left): here, effect sizes $\alpha\approx 0$, and fits are not significant (\Cref{tab:linear_fitting_D_PCA_twonn_prompts}). Results hold for The Pile, see \Cref{app:pile} for details.

\begin{figure*}
\floatbox[{\capbeside\thisfloatsetup{capbesideposition={left,top},capbesidewidth=0.35\textwidth}}]{figure}[\FBwidth]
{\caption{\textbf{$I_d$ tracks task performance}. \textbf{Top:} Layerwise $I_d$ over pre-training for Pythia-410m, 1.4b, and 6.9b. The phase transition of $I_d$ at $t\approx10^3$ holds across model size. \textbf{Bottom:} Zero-shot task performance across pre-training. Also at $t\approx10^3$, linguistic competence measured by task performance starts to increase for all LMs.}}
{\label{fig:perf-id} \includegraphics[width=0.6\textwidth]{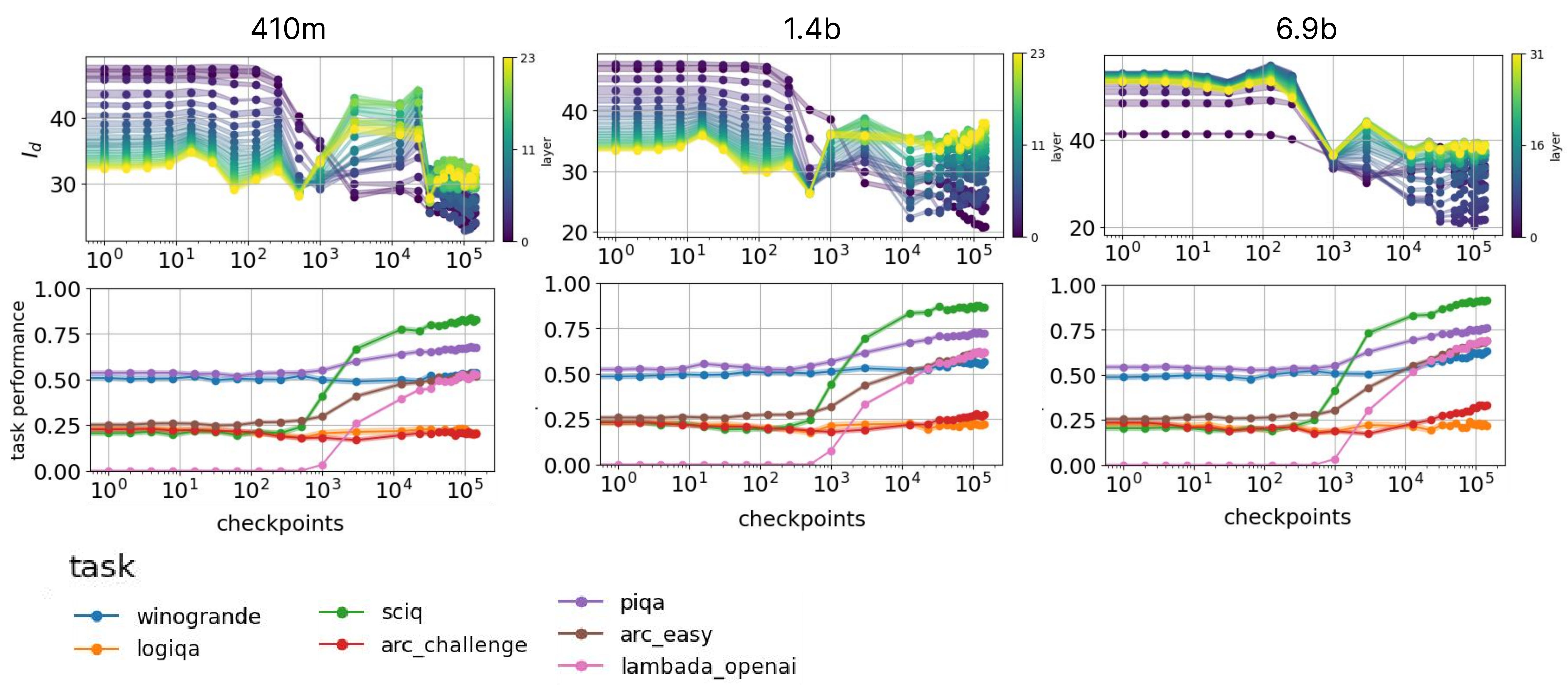}}
\end{figure*}

These results highlight key differences in how linear and nonlinear dimensions are recruited: LMs \emph{globally} distribute representations to occupy $d \propto D$ linear dimensions of the space, but their shape is \emph{locally} constrained to a low-dimensional ($I_d$) manifold. Robustness of $I_d$ to model size ($D$) suggests that LMs, once sufficiently performant, recover the degrees of freedom, or compositional complexity, underlying their inputs. 


\subsection{Feature complexity tracks
emergent linguistic abilities over training}
\label{sec:phase_transition}
Complex linguistic abilities require, by definition, a compositional understanding of language. We use linguistic task performance \citep{biderman2023pythia} as a cue for compositional understanding, finding that over training, models' feature complexity closely tracks the onset of linguistic capabilities. 

\Cref{fig:perf-id} shows the evolution of $I_d$ on the $k$=$1$ dataset (top), where each curve is one layer, with the evolution of LM benchmark performance (bottom), where each curve plots performance on an individual task. For all LMs, the evolution of $I_d$ closely tracks a sudden transition in task performance. In \Cref{fig:perf-id} top, we first observe that $I_d$ decreases sharply before checkpoint $10^3$ and then redistributes. At the same time $t\approx 10^3$, task performance rapidly improves (\Cref{fig:perf-id} bottom). Feature complexity evolution on The Pile is shown in \Cref{fig:pile_id_change_appendix}, exhibiting a similar transition to that shown in \Cref{fig:perf-id}. The existence of the phase transition in representational geometry $t\approx 10^3$ is robust to the dimensionality measure, and whether the data are shuffled, see \Cref{fig:id_change_appendix}. Our results resonate with \citet{chen2024sudden}, who observed in BERT a similar $I_d$ transition on the training corpus that coincided with the onset of higher-order linguistic capabilities.
Crucially, we show that the phase transition in feature geometry exists for natural language inputs \emph{beyond} in-distribution data, which was the subject of \citeauthor{chen2024sudden}, and, further, beyond grammatical data (\Cref{fig:id_change_appendix}) as a more general property of LM processing. Taken together, results show that feature complexity can signify when LMs gain linguistic abilities that require, by definition, compositional understanding.
 
\subsection{Feature complexity reflects input compositionality}
\label{sec:id_interpretation}
We just established that feature complexity is informative of when models gain linguistic behaviors that require compositional understanding. Now, we establish our key result, which is that, no matter the model or grammar, \textbf{feature complexity encodes input compositionality}. We first show that this holds for fully-trained models that have reached final linguistic competency. Then, using evidence from the LM's training phase, we show that the correspondence between feature complexity and input compositionality is present first as an inductive bias of the architecture that encodes superficial input complexity; but then, that it persists due to learned features that encode compositional meaning.

\begin{figure}
    \centering
    \includegraphics[width=\textwidth]{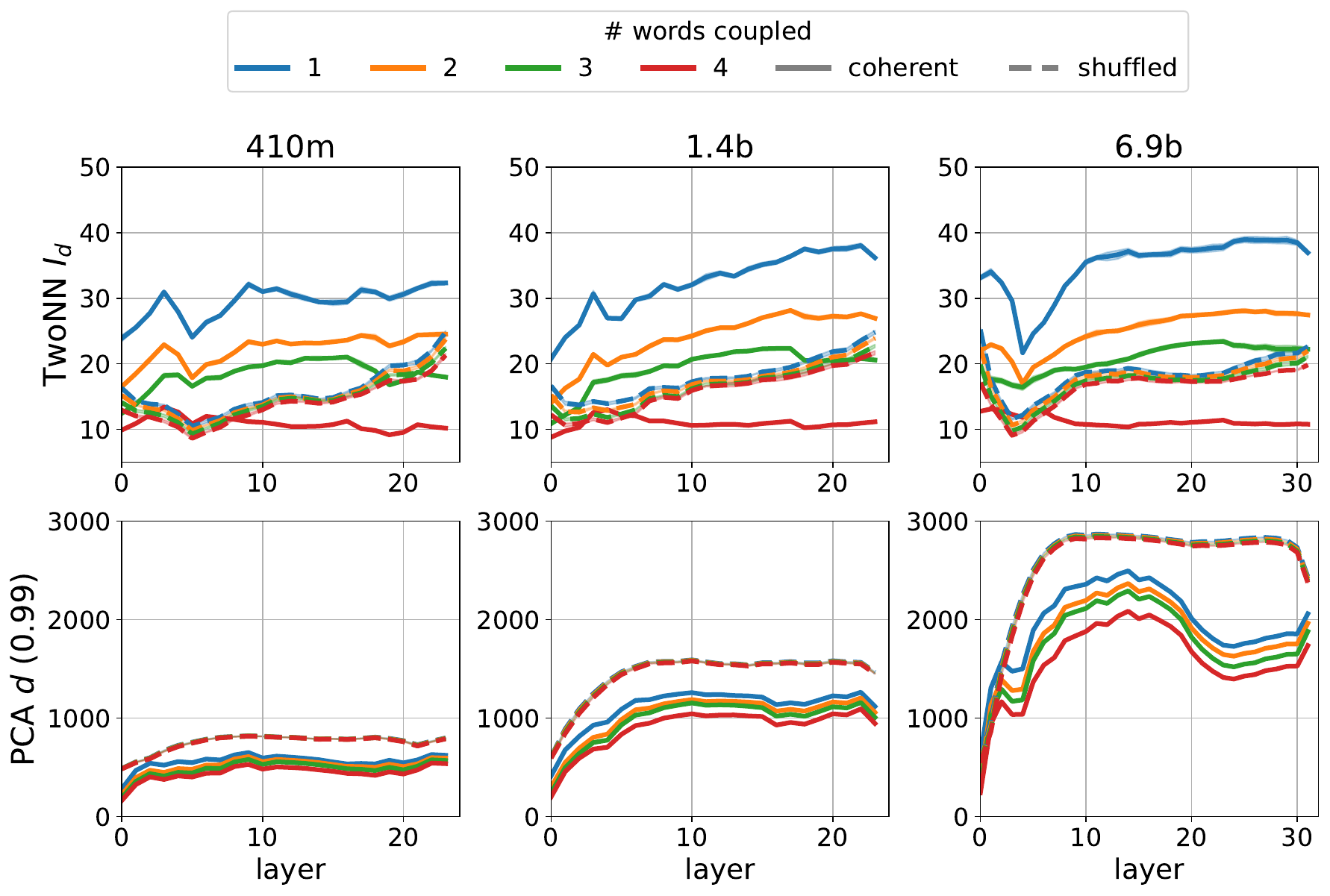}
    \caption{\textbf{Dimensionality over layers.} $I_d$ (top) and $d$ (bottom) over layers for Pythia 410m, 1.4b, 6.9b (left to right). Each color is a coupling factor $k\in 1\cdots 4$. Solid lines denote coherent, and dotted curves denote shuffled, text. For all LMs, lower $k$ implies higher $I_d$ and $d$ in both coherent and shuffled cases. For all LMs, shuffling collapses the $I_d$ but raises $d$. Curves averaged over 5 splits $\pm1$ SD (shaded, SDs very small).}
    \label{fig:pretrained_model_panel_id}
\end{figure}

\paragraph{Compositional complexity varying $k$}
On all fully-trained models and all datasets, representational dimensionality preserves relative dataset compositionality. \Cref{fig:pretrained_model_panel_id} shows for fully-trained Pythia 410m, 1.4b, and 6.9b that $I_d$ and $d$ increase with the compositionality of coherent inputs (solid curves): the highest curves (blue) correspond to the $k=1$ dataset, or $12$ degrees of freedom, and the lowest (red) denote the $k=4$ dataset, or $3$ degrees of freedom. The relative order of feature complexity, moreover, holds for all layers, seen by non-overlapping solid curves in \Cref{fig:pretrained_model_panel_id}.

Grammaticality is not a precondition for representational dimensionality to reflect superficial complexity of the data: in \Cref{fig:pretrained_model_panel_id} (top), dashed curves corresponding to shuffled text are also ordered $k=1\cdots 4$ top to bottom. While the relative order of superficial complexity is preserved in the LM's feature complexity for both grammatical and agrammatical datasets, the separation is greater for grammatical text (solid curves). We hypothesize that this is due to shuffled text being out-of-distribution, such that the model cannot integrate the sequences' meaning, but nevertheless preserve superficial complexity in its representations. This tendency holds for pre-trained LMs of all sizes and families (see \Cref{fig:all_results,fig:other_models_results}) and for different-length sentences (see \Cref{fig:huge_plot}).

\begin{figure}
    \centering
    \includegraphics[width=1\columnwidth]{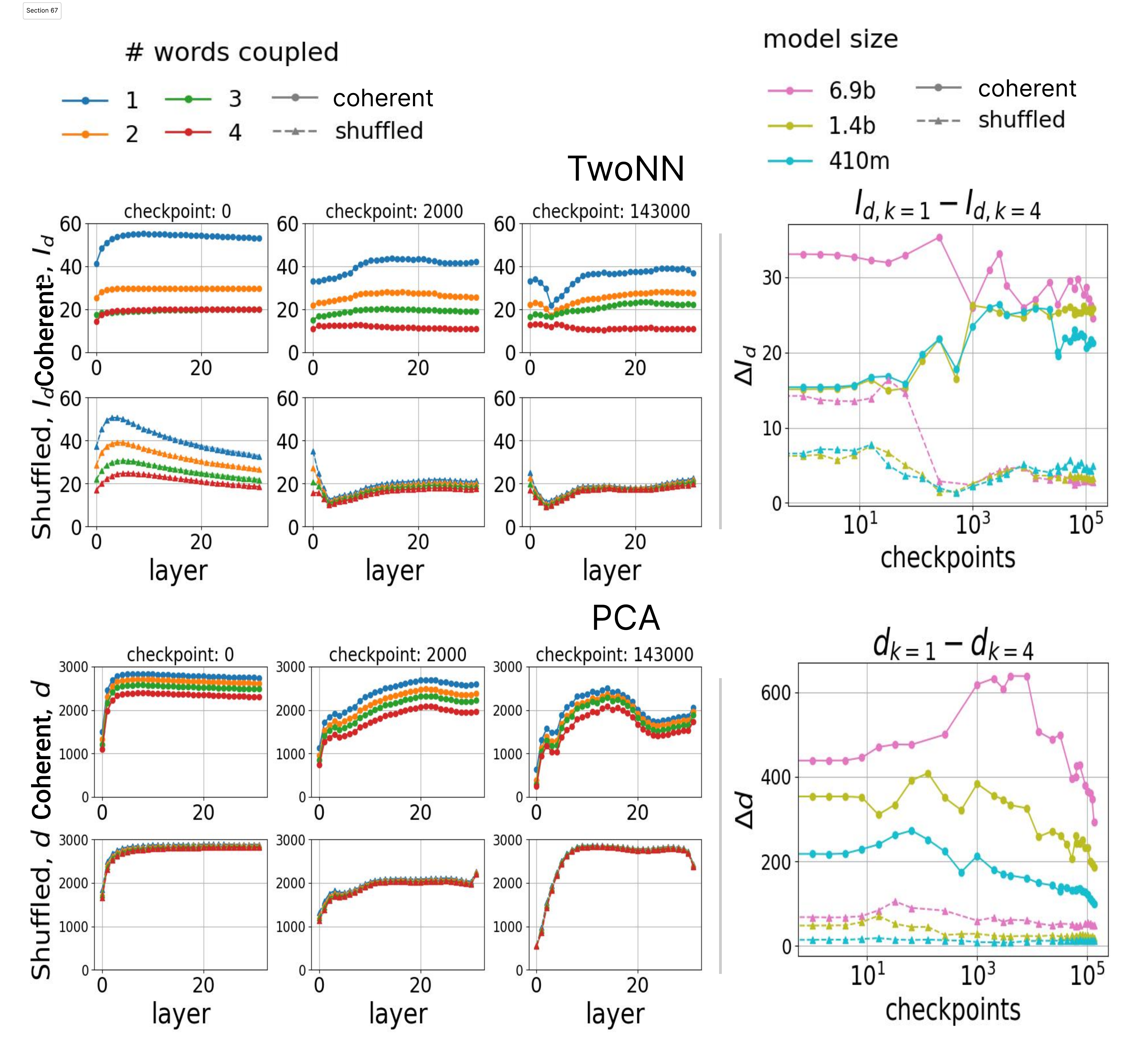}
    \caption{\textbf{Training dynamics of dimensionality.} (\textbf{Top}) TwoNN $I_d$; (\textbf{Bottom}) PCA $d$. \textbf{(Left): } For Pythia-6.9b, layerwise $I_d$ for $3$ training checkpoints for coherent vs.~shuffled sequences, with different coupling $k$. The $I_d$ difference on shuffled data with varying $k$ diminishes as training persists. All curves are shown with $\pm 1$ SD (SDs are very small). \textbf{(Right):} $\Delta I_d$ between $k=1, k=4$ across training for various model sizes (different colors).}
    \label{fig:id_change_coherent_shuffle}
\end{figure}

The relationship between dimensionality and superficial complexity, controlled by $k$, for coherent text is \emph{not} an emergent feature over training. In \Cref{fig:id_change_coherent_shuffle} (left), the inverse relationship between $k$ and both $I_d$ and $d$ is present throughout training. But, the reason for this relation differs at the start and end: in shuffled text, where phrase-level semantics is ablated, the inverse relationship between $k$ and dimensionality exists at the \emph{start} and greatly diminishes by the end of training, while in coherent text it is salient throughout. Together, results show an inductive bias of the LM's architecture to preserve input complexity in its representations. Then, over training, differences in dimensionality may be increasingly explained by features beyond surface variation of inputs. We claim these features are semantic and provide evidence in what follows.

\paragraph{Breaking compositional structure} Shuffling sequences destroys their meaning, removing dataset complexity attributed to sentence-level semantics. \Cref{fig:pretrained_model_panel_id} shows, for fully-trained models, dimensionality over layers for coherent and shuffled inputs. Here, nonlinear $I_d$ and linear $d$ show opposing trends: $I_d$ for shuffled text \emph{collapses} to a low range, while $d$ \emph{increases}, seen by the dashed curves in each plot compared to solid curves. This trend is robust to sentence length (\Cref{fig:huge_plot}) and model size (\Cref{fig:all_results}) and family (\Cref{fig:other_models_results}).

We refer to the phenomenon where shuffling destroys phrase-level semantics and $I_d$, also attested for naturalistic sentences \citep{cheng2025emergence}, as \emph{shuffling feature collapse}. Evidence from training dynamics further suggests that this feature collapse is due to semantics. We saw in \Cref{sec:phase_transition} that training step $t=10^3$ approximately marked a phase transition after which the LM's linguistic competencies sharply rose. Crucially, the epoch $t=10^3$ preceding the sharp increase in linguistic capabilities is also the first to exhibit shuffling feature collapse. \Cref{fig:id_change_coherent_shuffle} (right) shows the $\Delta I_d$ between the $k=1$ and $k=4$ dataset for several model sizes, across training (x-axis). Shuffling feature collapse, given by low $\Delta I_d$, occurs around $t=10^3$ for all models. On the other hand, $\Delta I_d$ for coherent text stabilizes to around $25$ for different model sizes. This transition does not occur for linear $d$, see \Cref{fig:id_change_coherent_shuffle} (right, bottom). This suggests that shuffling feature collapse for $I_d$ is symptomatic of when the LM learns to extract meaningful semantic features. We investigate why shuffling feature collapse exists for $I_d$, but not for $d$, in the following section. For how shuffling feature collapse emerges with sequence length, see \Cref{app:sequence_length}.

\begin{table*}[h!!]
\centering 
\scalebox{1.0}{
\begin{tabular}{lrrrrrrrrr}
\toprule
Spearman $\rho$ & 14m & 70m & 160m & 410m & 1.4b & 6.9b & 12b & Llama & Mistral \\
\midrule
$I_d$ & -0.20 & -0.06 & -0.20 & -0.05 & 0.04 & 0.01 & 0.05 & -0.36 & 0.00 \\
$d$ & 0.90* & 0.47\textsuperscript{\textdagger} & -0.50\textsuperscript{\textdagger} & 0.96* & 0.96* & 0.92* & 0.86* & 1.0* & 1.0* \\
\bottomrule
\end{tabular}}
\caption{\textbf{Spearman correlations between dimensionality and estimated Kolmogorov complexity}. Spearman correlations $\rho$ between the 
\texttt{gzip}ped dataset size (KB) and representational dimensionality (rows), averaged over layers, for all \textbf{Pythia} sizes (left 7 columns) and \textbf{Llama} and \textbf{Mistral} (right 2 columns). Values with * are significant at a p-value threshold of $0.05$, and \textdagger \space at $0.1$. Across LMs, average-layer $I_d$ is not correlated to the estimated KC, or surface complexity, of datasets. Linear $d$ is highly correlated to estimated KC, except the outlier 160m. \\ }
\label{tab:gzip}
\end{table*}


\section{Form-meaning dichotomy in representation learning}
We interpret shuffling feature collapse using an argument from \citet{Recanatesi_Farrell_Lajoie_Deneve_Rigotti_Shea-Brown_2021}, who propose that predictive coding (i.e. the theory that the goal of any intelligent agent is to minimize prediction errors between what it expects and what is actually perceived) requires a model to satisfy two objectives: \textcolor{MidnightBlue}{(1)} encode the vast ``input and output space", exerting upward pressure on feature complexity, while \textcolor{BrickRed}{(2)} extracting latent features to support prediction, exerting a downward pressure on complexity. \citet{Recanatesi_Farrell_Lajoie_Deneve_Rigotti_Shea-Brown_2021} argue that pressure \textcolor{MidnightBlue}{(1)} expands the linear representation space ($d$), while \textcolor{BrickRed}{(2)} compresses representations to a $I_d$-dimensional manifold. Recall that we quantify a dataset's superficial complexity by its KC. We saw that shuffling words, which increases inputs' KC, also increases $d$, but that it destroys compositional semantics, collapsing $I_d$ (\Cref{fig:pretrained_model_panel_id}). These results suggest a dichotomy in what linear and nonlinear complexity encode, the former capturing superficial complexity \textcolor{MidnightBlue}{(form)}, and the latter, latent compositional structure \textcolor{BrickRed}{(meaning)}.

We hypothesize $d$ to capture surface variation, and $I_d$ to encode meaningful compositional variation. We saw the latter in the previous section: $I_d$ collapses to a narrow range in the absence of compositional semantics while $d$ does not, suggesting $I_d$, not $d$, encodes compositionality. Now, we show that $d$, not $I_d$, encodes superficial complexity. \Cref{tab:gzip} shows Spearman correlations $\rho$ between KC (estimated with \texttt{gzip}) and dimensionality. For all LMs, \texttt{gzip} highly correlates to average layerwise $d$ but not to $I_d$; we discuss the outlier Pythia-160m in \Cref{app:kc}. The high $\rho(d, \texttt{gzip})$ is surprisingly consistent across layers (\Cref{fig:layer-correlations,fig:other_models_layer_correlations}) and already present as an inductive bias of the architecture (\Cref{fig:time-evolution-kc}, \Cref{app:kc} for training discussion). Instead, the correlation to $I_d$ is seldom significant for all of training. This suggests surface complexity, already present in the LM's inputs, is \emph{preserved} by the architecture, while meaning complexity is instead \emph{learned} over training.

\section{Discussion}
On extensive experiments spanning degree of compositionality, sequence length, model size and training time, we found strong links between the compositionality of linguistic inputs and the complexity of their representations. We asked what aspects of feature complexity encode inputs' \textcolor{MidnightBlue}{(1) surface variation} and \textcolor{BrickRed}{(2) compositional structure}, finding
that linear $d$ tracks \textcolor{MidnightBlue}{(1)} and nonlinear $I_d$ tracks \textcolor{BrickRed}{(2)}. We showed feature complexity to encode inputs' superficial complexity as an inductive bias of the architecture; but that, over training, the bias is shed to reflect meaningful linguistic compositionality. This change marks an onset in linguistic competence, a cue for compositional understanding.

A central throughline in our results is that LMs represent their inputs on low-dimensional nonlinear manifolds, yet, at the same time, in high-dimensional linear subspaces. This tendency suggests a solution to the curse of dimensionality that also enjoys its blessings. High-dimensional representations classically engender overfitting and poor generalization~\citep{curse_dim}; but, they may form low-dimensional manifolds that capture the task's latent, in our case, \emph{compositional}, structure~\citep{de_nonlinear_2023}. At the same time, higher linear dimensionality has been shown to improve generalization thanks to its expressivity \citep{Cohen_Chung_Lee_Sompolinsky_2020,Elmoznino2023HighperformingNN,Sorscher_Ganguli_Sompolinsky_2022}. Intriguingly, the dual patterning of intrinsic and effective dimension, where latent structure is captured by low-$I_d$ manifolds that live in high-$d$ linear spaces, has been observed in the neuroscience literature for both biological and artificial systems \citep{JAZAYERI2021113,Recanatesi_Farrell_Lajoie_Deneve_Rigotti_Shea-Brown_2021,Haxby2011ACH,Huth2012ACS,de_nonlinear_2023,Manley2024SimultaneousCD}. That LMs also conform to this picture suggests that representation spaces that are both highly expressive ($d$) yet geometrically constrained ($I_d$) may be desirable in intelligent systems that recover compositional structure from data.

\section*{Limitations}
\begin{itemize}
    \item While the present work is the first to show that nonlinear and linear representational dimension in LMs correspond to a form-meaning dichotomy, further work is needed to disentangle how nonlinear and linear features causally contribute to predictive coding. This would require careful probing and feature attribution.
    \item Due to computational constraints, the present work is limited to several syntactic structures. Future work can apply this framework to investigate other structures like recursive embedding. 
    \item Also due to computational constraints, the present work is limited to model sizes of up to 8B parameters. We encourage further, better-resourced research to apply our framework to larger models. 
    \item Dimensionality measures tell us how complex the features are, but not \emph{what} they are. How to isolate and interpret nonlinear features remains an open problem in the literature.
\end{itemize}

\subsubsection*{Acknowledgements}
This project has received funding from the European Research Council (ERC) under the European Union’s Horizon 2020 research and innovation programme (grant agreement No. 101019291). This paper reflects the authors’ view only, and the funding agency is not responsible for any use that may be made of the information it contains. JHL thanks the Gatsby Charitable Foundation (GAT3755) and The Wellcome Trust (219627/Z/19/Z). TJ is funded by FRQNT. EC, JHL and TJ thank the Center for Brain, Minds and Machines for hosting the collaboration.

The authors thank Marco Baroni, Noga Zaslavsky, Erin Grant, Francesca Franzon, Alessandro Laio, Andrew Saxe, and members of the COLT group at Universitat Pompeu Fabra for valuable feedback.

\subsubsection*{Author Contributions}
EC and LY contributed the idea of exploring ID, compression, and linguistic compositionality in the static setting. JHL greatly expanded this idea to incorporate investigation of dynamics and evolution of the features of interest over training. EC, JHL and TJ created the grammar, implemented and ran experiments; TJ led the static, and JHL led the dynamic, comparison of geometry to compositionality. EC helped with computation, did Kolmogorov complexity experiments with TJ, and led the writing. All authors contributed to the manuscript.

\bibliography{custom}

\appendix
\renewcommand\thefigure{\thesection.\arabic{figure}}    
\renewcommand\thetable{\thesection.\arabic{table}}
\renewcommand\theequation{\thesection.\arabic{equation}}
\section{Computing resources}
\label{app:compute}

All experiments were run on a cluster with 12 nodes with 5 NVIDIA A30 GPUs and 48 CPUs each. Extracting LM representations took a few wall-clock hours per model-dataset computation. $I_d$ computation took approximately 0.5 hours per model-dataset computation. Taking parallelization into account, we estimate the overall wall-clock time taken by all experiments, including failed runs, preliminary experiments, etc., to be of about 10 days.

\section{Assets}
\label{app:assets}

\begin{description}
    \item[Pile] \url{https://huggingface.co/datasets/NeelNanda/pile-10k}; license: bigscience-bloom-rail-1.0
    \item[Pythia] \url{https://huggingface.co/EleutherAI/pythia-6.9b-deduped}; license: apache-2.0
    \item[scikit-dimension] \url{https://scikit-dimension.readthedocs.io/en/latest/}; license: bsd-3-clause
    \item[PyTorch] \url{https://scikit-learn.org/}; license: bsd
\end{description}

\section{Intrinsic Dimension Details}
\label{app:id}
\setcounter{figure}{0}    
\setcounter{table}{0} 
$I_d$ estimation methods practically rely on a finite set of points and their nearest-neighbor structure in order to compute an estimated dimensionality value. The underlying geometric calculations assume that these are points sampled from a continuum, such as a lower-dimensional non-linear manifold. In our case, we actually have a discrete set of points so the notion of an underlying manifold is not strictly applicable. However, we can ask the question: if those points had been sampled from a manifold, what would the estimated $I_d$ be? Since the algorithms themselves only require a discrete set of points, they can be used to answer that question.

\section{Other Dimensionality Estimators}
\label{app:id_estimators}


\paragraph{Maximum Likelihood Estimator} In addition to TwoNN, we considered \citet{Levina_Bickel_2004}'s Maximum Likelihood Estimator (MLE), a similar, nonlinear measure of $I_d$. MLE has been used in prior works on representational geometry such as \citep{cai2021isotropy,Cheng:etal:2023,pope2021the}, and similarly models the number of points in a neighborhood around a reference point $x$ to follow a Poisson point process. For details we refer to the original paper \citep{Levina_Bickel_2004}. Like past work \citep{Facco_d'Errico_Rodriguez_Laio_2017,Cheng:etal:2023}, we found MLE and TwoNN to be highly correlated, producing results that were nearly identical: compare \Cref{fig:id_size_scaling} left to \Cref{fig:pr_mle} left, and \Cref{fig:all_results} top to \Cref{fig:mle_pr_over_layers} top.

\paragraph{Participation Ratio} For our primary linear measure of dimensionality $d$, we computed PCA and took the number of components that explain 99\% of the variance. In addition to PCA, we computed the Participation Ratio (PR), defined as $(\sum_i \lambda_i)^2 / (\sum_i \lambda_i ^2)$ \citep{Gao2017ATO}. We found PR to give results that were incongruous with intuitions about linear dimensionality. In particular, it produced a lower dimensionality estimate than the nonlinear estimators we tested; see, e.g., \Cref{fig:pr_mle}, where the PR-$d$ for coherent text is less than that of TwoNN. This contradicts the mathematical relationship that $I_d \leq d \leq D$. This may be because, empirically, PR-$d$ corresponded to explained variances of $60-80\%$, which are inadequate to describe the bounding linear subspace for the representation manifold. Therefore, while we report the mean PR-$d$ over model size in \Cref{fig:pr_mle} and the dimensionality over layers in \Cref{fig:mle_pr_over_layers} for completeness, we do not attempt to interpret them.

\section{Controlled Grammar}
\label{app:grammar}
\setcounter{figure}{0}    
\setcounter{table}{0} 
\begin{figure*}
    \centering
    \includegraphics[width=\linewidth]{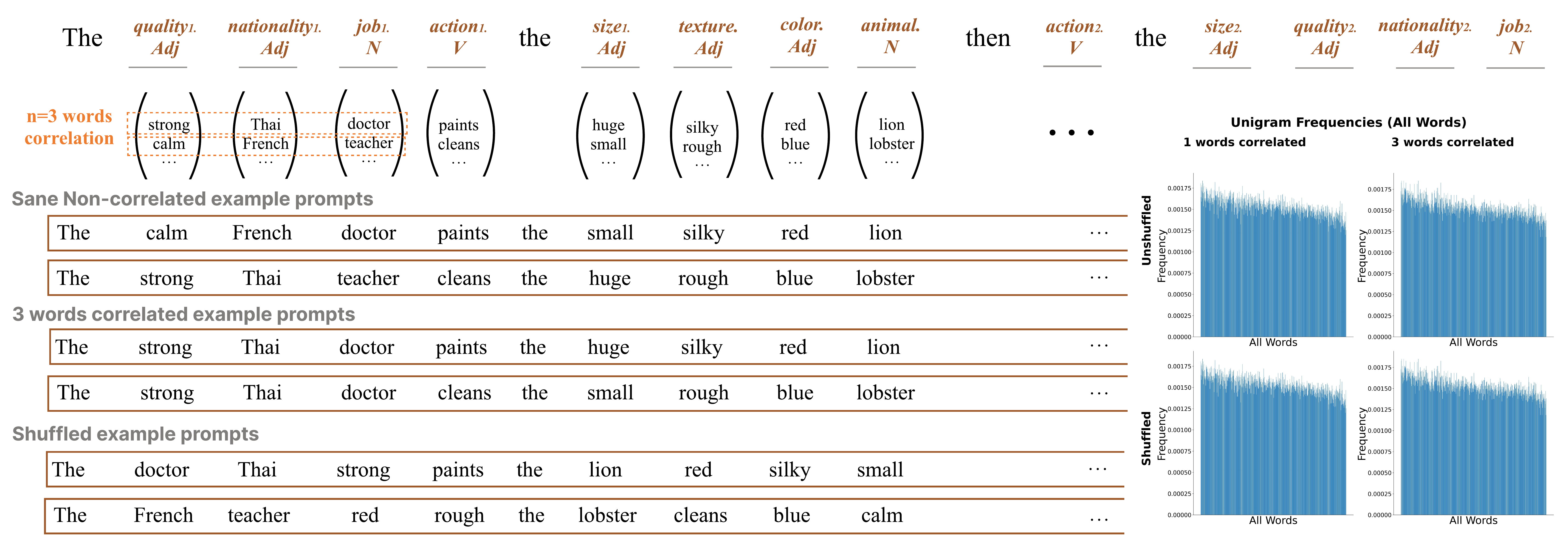}
    \caption{\textbf{Dataset structure and distributional properties.} \textbf{Top:} The structure of the stimulus dataset. The top row shows the ordering of word categories, such as quality.ADJ or animal.N; below it, the vocabulary for each category, including words like ``strong" (quality.ADJ) and ``lion" (animal.N), respectively. When controlling the degree of dataset compositionality, contiguous word positions are coupled. For instance, when $k=3$, the first vocabulary indices for quality$_1$.ADJ, nationality$_1$.ADJ, and job$_1$.N are tied together, such that ``strong Thai doctor" or ``calm French teacher" can be sampled, but ``strong French doctor" cannot. \textbf{Left:} Examples of generated prompts for the normal, $k=3$, and shuffled settings. \textbf{Right:} When controlling the compositionality across $k=1\cdots 4$, word unigram frequencies are preserved in the resulting datasets, shown in the distributions looking identical.}
    \label{fig:prompt}
\end{figure*}

Here, we describe the sampling procedure of the controlled grammar in more detail.

\subsection{Sampling Procedure}
Let there be $l$ variable words in the sequence (these are the colored words in the grammar in Section 3.1). Each position $i = 1\cdots l$ is associated with a vocabulary $\mathcal V^{(i)}$ which contains vocabulary items $v_j^{(i)}$, $j=1\cdots |\mathcal V^{(i)}|$. We set $|\mathcal V^{(i)}|=50$ for all $i$. All words in $\mathcal V^{(i)}$ are listed in the below Table. 

We state the sampling procedure, illustrated in \Cref{fig:prompt}, as follows.

\begin{itemize}
\item $\mathbf{k=1}$ For each position $i$ in the sentence, sample the word $w_{i} \sim \text{Unif}(\mathcal V^{(i)})$.
\item $\mathbf{k=2}$ Couples the vocabularies over bigrams in the sentence. Let $\circ$ be a concatenation operator. We construct a new \emph{coupled} vocabulary $\mathcal V^{(i,i+1)}$ consisting of bigrams $v_j^{(i,i+1)} \triangleq v_j^{(i)}\circ v_j^{(i+1)}$, $j=1\cdots |\mathcal V^{(i)}|$. Then, for $i=1, 3, 5, \cdots l-1$, sample bigrams $w_{i}\circ w_{i+1} \sim \text{Unif}({\mathcal V^{(i,i+1)}})$. 
\item $\mathbf{k=3}$ Couples trigrams in the sentence. \\ Similarly, a new vocabulary over trigrams is constructed: $\mathcal V^{(i,i+1,i+2)}$ consists of trigrams $v_j^{(i,i+1,i+2)} := v_j^{(i)}\circ v_j^{(i+1)}\circ v_j^{(i+2)}$. For $i=1, 4, 7, \cdots l-2$, sample trigrams $w_{i}w_{i+1}w_{i+2} \sim \text{Unif}({\mathcal V^{(i,i+1,i+2)}})$...
\end{itemize}

and so on. 

Although the LM is likely exposed to aspects of the syntactic structure and vocabulary items during training, primitives are sampled independently without considering their probability in relationship to others in the sentence. Therefore, generated sentences are highly unlikely to be in the training data, thus memorized by the LM.\footnote{\noindent We can't verify utterances aren't in the training set, as at the time of submission, it is not possible to search The Pile.} 

We design 5 different grammars of varying lengths $l \in \{5, 8, 11, 15, 17\}$ words.. The $17$-word grammar is the one used for all controlled grammar experiments except the "Varying Sequence Length" experiments (\Cref{app:sequence_length}). The structures of the grammars can be found below.

\subsection{Length: 5 words}
\begin{center}
\parbox{\columnwidth}{%
The \texttt{\textcolor{olive}{[job$_1$.N]} \textcolor{Periwinkle}{[action$_1$.V]}} the \texttt{\textcolor{olive}{[animal.N]}.}
}%
\end{center}
\subsection{Length: 8 words}
\begin{center}
\parbox{\columnwidth}{%
The \texttt{\textcolor{olive}{[nationality$_1$.ADJ] [job$_1$.N]} \textcolor{Periwinkle}{[action$_1$.V]}} the \texttt{\textcolor{olive}{[color.ADJ] [texture.ADJ] [animal.N]} }
}%
\end{center}
\subsection{Length: 11 words}
\begin{center}
\parbox{\columnwidth}{%
The \texttt{\textcolor{olive}{[size$_2$.ADJ] [quality$_1$.ADJ] [nationality$_1$.ADJ] [job$_1$.N]} \textcolor{Periwinkle}{[action$_1$.V]}} the \texttt{\textcolor{olive}{[size$_1$.ADJ] [color.ADJ] [texture.ADJ] [animal.N]}}
}%
\end{center}
\subsection{Length: 15 words}
\begin{center}
\parbox{\columnwidth}{%
The \texttt{\textcolor{olive}{[quality$_1$.ADJ] [nationality$_1$.ADJ] [job$_1$.N]} \textcolor{Periwinkle}{[action$_1$.V]}} the \texttt{\textcolor{olive}{[size$_1$.ADJ] [color.ADJ] [texture.ADJ] [animal.N]}} then \texttt{\textcolor{Periwinkle}{[action$_2$.V]}} the \texttt{\textcolor{olive}{[size$_2$.ADJ][job$_2$.N]}.}
}%
\end{center}
\subsection{Length: 17 words}

\begin{center}
\parbox{\columnwidth}{%
The \texttt{\textcolor{olive}{[quality$_1$.ADJ] [nationality$_1$.ADJ] [job$_1$.N]} \textcolor{Periwinkle}{[action$_1$.V]}} the \texttt{\textcolor{olive}{[size$_1$.ADJ] [texture.ADJ] [color.ADJ] [animal.N]}} then \texttt{\textcolor{Periwinkle}{[action$_2$.V]}} the \texttt{\textcolor{olive}{[size$_2$.ADJ] [quality$_2$.ADJ][nationality$_2$.ADJ][job$_2$.N]}.}
}%
\end{center}

Each category, colored and enclosed in brackets, is sampled from a vocabulary of 50 possible words, listed in the table on the following page. 

The 50 words for each category were selected by prompting GPT4 with appropriate prompts. For example, "Please generate 50 nouns referring to jobs" or “Please generate 50 verbs referring to actions that a human can do to an animal”.
When there were 2 sets of words for a word category (for example action$_1$ and action$_2$), we provided GPT4 with the previous 50 words and prompted it appropriately. For example, "Please generate 50 different verbs referring to actions that a human can do to another human." When GPT4 mistakenly generated repeat words, we prompted it to replace the repeat words with other appropriate words."
\clearpage
\onecolumn
\begin{longtable}{|p{0.3\linewidth}|p{0.65\linewidth}|}
\hline
\textbf{Category} & \textbf{Words} \\
\hline
job$_1$ & teacher, doctor, engineer, chef, lawyer, plumber, electrician, accountant, nurse, mechanic, architect, dentist, programmer, photographer, painter, firefighter, police, pilot, farmer, waiter, scientist, actor, musician, writer, athlete, designer, carpenter, librarian, journalist, psychologist, gardener, baker, butcher, tailor, cashier, barber, janitor, receptionist, salesperson, manager, tutor, coach, translator, veterinarian, pharmacist, therapist, driver, bartender, security, clerk \\
\hline
job$_2$ & banker, realtor, consultant, therapist, optometrist, astronomer, biologist, geologist, archaeologist, anthropologist, economist, sociologist, historian, philosopher, linguist, meteorologist, zoologist, botanist, chemist, physicist, mathematician, statistician, surveyor, pilot, steward, dispatcher, ichthyologist, oceanographer, ecologist, geneticist, microbiologist, neurologist, cardiologist, pediatrician, surgeon, anesthesiologist, radiologist, dermatologist, gynecologist, urologist, psychiatrist, physiotherapist, chiropractor, nutritionist, personal trainer, yoga instructor, masseur, acupuncturist, paramedic, midwife \\
\hline

animal & dog, cat, elephant, lion, tiger, giraffe, zebra, monkey, gorilla, chimpanzee, bear, wolf, fox, deer, moose, rabbit, squirrel, raccoon, beaver, otter, penguin, eagle, hawk, owl, parrot, flamingo, ostrich, peacock, swan, duck, frog, toad, snake, lizard, turtle, crocodile, alligator, shark, whale, dolphin, octopus, jellyfish, starfish, crab, lobster, butterfly, bee, ant, spider, scorpion \\
\hline

color & red, blue, green, yellow, purple, orange, pink, brown, gray, black, white, cyan, magenta, turquoise, indigo, violet, maroon, navy, olive, teal, lime, aqua, coral, crimson, fuchsia, gold, silver, bronze, beige, tan, khaki, lavender, plum, periwinkle, mauve, chartreuse, azure, mint, sage, ivory, salmon, peach, apricot, mustard, rust, burgundy, mahogany, chestnut, sienna, ochre \\
\hline

size$_1$ & big, small, large, tiny, huge, giant, massive, microscopic, enormous, colossal, miniature, petite, compact, spacious, vast, wide, narrow, slim, thick, thin, broad, expansive, extensive, substantial, boundless, considerable, immense, mammoth, towering, titanic, gargantuan, diminutive, minuscule, minute, hulking, bulky, hefty, voluminous, capacious, roomy, cramped, confined, restricted, limited, oversized, undersized, full, empty, half, partial \\
\hline

size$_2$ & lengthy, short, tall, long, deep, shallow, high, low, medium, average, moderate, middling, intermediate, standard, regular, normal, ordinary, sizable, generous, abundant, plentiful, copious, meager, scanty, skimpy, inadequate, sufficient, ample, excessive, extravagant, exorbitant, modest, humble, grand, majestic, imposing, commanding, dwarfed, diminished, reduced, enlarged, magnified, amplified, expanded, contracted, shrunken, swollen, bloated, inflated, deflated \\
\hline

nationality$_1$ & American, British, Canadian, Australian, German, French, Italian, Spanish, Japanese, Chinese, Indian, Russian, Brazilian, Mexican, Argentinian, Turkish, Egyptian, Nigerian, Kenyan, African, Swedish, Norwegian, Danish, Finnish, Icelandic, Dutch, Belgian, Swiss, Austrian, Greek, Polish, Hungarian, Czech, Slovak, Romanian, Bulgarian, Serbian, Croatian, Slovenian, Ukrainian, Belarusian, Estonian, Latvian, Lithuanian, Irish, Scottish, Welsh, Portuguese, Moroccan, Algerian \\
\hline

nationality$_2$ & Vietnamese, Thai, Malaysian, Indonesian, Filipino, Singaporean, Nepalese, Bangladeshi, Maldivian, Pakistani, Afghan, Iranian, Iraqi, Syrian, Lebanese, Israeli, Saudi, Emirati, Qatari, Kuwaiti, Omani, Yemeni, Jordanian, Palestinian, Bahraini, Tunisian, Libyan, Sudanese, Ethiopian, Somali, Ghanaian, Ivorian, Senegalese, Malian, Cameroonian, Congolese, Ugandan, Rwandan, Tanzanian, Mozambican, Zambian, Zimbabwean, Namibian, Botswanan, New Zealander, Fijian, Samoan, Tongan, Papuan, Marshallese \\
\hline

action$_1$ & feeds, walks, grooms, pets, trains, rides, tames, leashes, bathes, brushes, adopts, rescues, shelters, houses, cages, releases, frees, observes, studies, examines, photographs, films, sketches, paints, draws, catches, hunts, traps, chases, pursues, tracks, follows, herds, corrals, milks, shears, breeds, mates, clones, dissects, stuffs, mounts, taxidermies, domesticates, harnesses, saddles, muzzles, tags, chips, vaccinates \\
\hline

action$_2$ & hugs, kisses, loves, hates, admires, respects, befriends, distrusts, helps, hurts, teaches, learns from, mentors, guides, counsels, advises, supports, undermines, praises, criticizes, compliments, insults, congratulates, consoles, comforts, irritates, annoys, amuses, entertains, bores, inspires, motivates, discourages, intimidates, impresses, disappoints, surprises, shocks, delights, disgusts, forgives, resents, envies, pities, understands, misunderstands, trusts, mistrusts, betrays, protects \\
\hline

quality$_1$ & good, bad, excellent, poor, superior, inferior, outstanding, mediocre, exceptional, sublime, superb, terrible, wonderful, awful, great, horrible, fantastic, dreadful, marvelous, atrocious, splendid, appalling, brilliant, dismal, fabulous, lousy, terrific, abysmal, incredible, substandard, amazing, disappointing, extraordinary, stellar, remarkable, unremarkable, impressive, unimpressive, admirable, despicable, praiseworthy, blameworthy, commendable, reprehensible, exemplary, subpar, ideal, flawed, perfect, imperfect \\
\hline

quality$_2$ & acceptable, unacceptable, satisfactory, unsatisfactory, sophisticated, insufficient, adequate, exquisite, suitable, unsuitable, appropriate, inappropriate, fitting, unfitting, proper, improper, correct, incorrect, right, wrong, accurate, inaccurate, precise, imprecise, exact, inexact, flawless, faulty, sound, unsound, reliable, unreliable, dependable, undependable, trustworthy, untrustworthy, authentic, fake, genuine, counterfeit, legitimate, illegitimate, valid, invalid, legal, illegal, ethical, unethical, moral, immoral \\
\hline

texture & smooth, rough, soft, hard, silky, coarse, fluffy, fuzzy, furry, hairy, bumpy, lumpy, grainy, gritty, sandy, slimy, slippery, sticky, tacky, greasy, oily, waxy, velvety, leathery, rubbery, spongy, springy, elastic, pliable, flexible, rigid, stiff, brittle, crumbly, flaky, crispy, crunchy, chewy, stringy, fibrous, porous, dense, heavy, light, airy, feathery, downy, woolly, nubby, textured \\
\hline
\end{longtable}

\twocolumn

\section{Benchmark tasks}
\label{app:tasks}
\setcounter{figure}{0}    
\setcounter{table}{0} 
Here we briefly summarize the benchmark tasks that we use to evaluate Pythia checkpoints as described in Section 4.3. In figure \Cref{fig:perf-id}, we did not include WSC (Winogrande Schema Challenge) which was originally included in \citeauthor{biderman2023pythia}, as it has been proposed that WSC dataset performance on LMs might be corrupted by spurious biases in the dataset \citep{sakaguchi2021winogrande}. Instead, we only presented the evaluation from WinoGrande task, which is inspired from original WSC task but adjusted to reduce the systematic bias \citep{sakaguchi2021winogrande}.

\paragraph{WinoGrande} WinoGrande \citep{sakaguchi2021winogrande} is a dataset designed to test commonsense reasoning by building on the structure of the Winograd Schema Challenge \citep{levesque2012winograd}. It presents sentence pairs with subtle ambiguities where understanding the correct answer requires world knowledge and commonsense reasoning. It challenges models to differentiate between two possible resolutions of pronouns or references, making it a benchmark for evaluating an AI’s ability to understand context and reasoning.

\paragraph{LogiQA} LogiQA \citep{liu2020logiqa} is an NLP benchmark for evaluating logical reasoning abilities in models. It consists of multiple-choice questions derived from logical reasoning exams for human students. The questions test various forms of logical reasoning, such as deduction, analogy, and quantitative reasoning, making it ideal for assessing how well AI can handle structured logical problems.

\paragraph{SciQ} SciQ \citep{welbl-etal-2017-crowdsourcing} is a dataset focused on scientific question answering, based on material from science textbooks. It features multiple-choice questions related to science topics like biology, chemistry, and physics. The benchmark is designed to test a model's ability to comprehend scientific information and answer questions using factual knowledge and reasoning.

\paragraph{ARC Challenge} The ARC (AI2 Reasoning Challenge) Challenge Set \citep{clark2018think} is a benchmark designed to test models on difficult, grade-school-level science questions. It presents multiple-choice questions that are challenging due to requiring complex reasoning, inference, and background knowledge beyond simple retrieval-based approaches. It is a tougher subset of the larger ARC dataset.

\paragraph{PIQA} PIQA (Physical Interaction QA) \citep{bisk2020piqa} is a benchmark designed to test models on physical commonsense reasoning. The questions require understanding basic physical interactions, like how objects interact or how everyday tasks are performed. It focuses on scenarios that involve intuitive knowledge of the physical world, making it a useful benchmark for evaluating practical commonsense in models.

\paragraph{ARC Easy} ARC Easy is the easier subset of the AI2 Reasoning Challenge, consisting of grade-school-level science questions that require less complex reasoning compared to the Challenge set. This benchmark is meant to evaluate models' ability to handle straightforward factual and retrieval-based questions, making it more accessible for baseline NLP models.

\paragraph{LAMBADA} LAMBADA \citep{paperno-etal-2016-lambada} is a reading comprehension benchmark where models must predict the last word of a passage. The challenge lies in the fact that understanding the entire context of the passage is necessary to guess the correct word. This benchmark tests a model’s long-range context comprehension and coherence skills in natural language.

\section{Additional Results: Controlled Grammar}
\setcounter{figure}{0}    
\setcounter{table}{0} 

\begin{figure*}[htp]
    \centering
    \includegraphics[width=\linewidth]{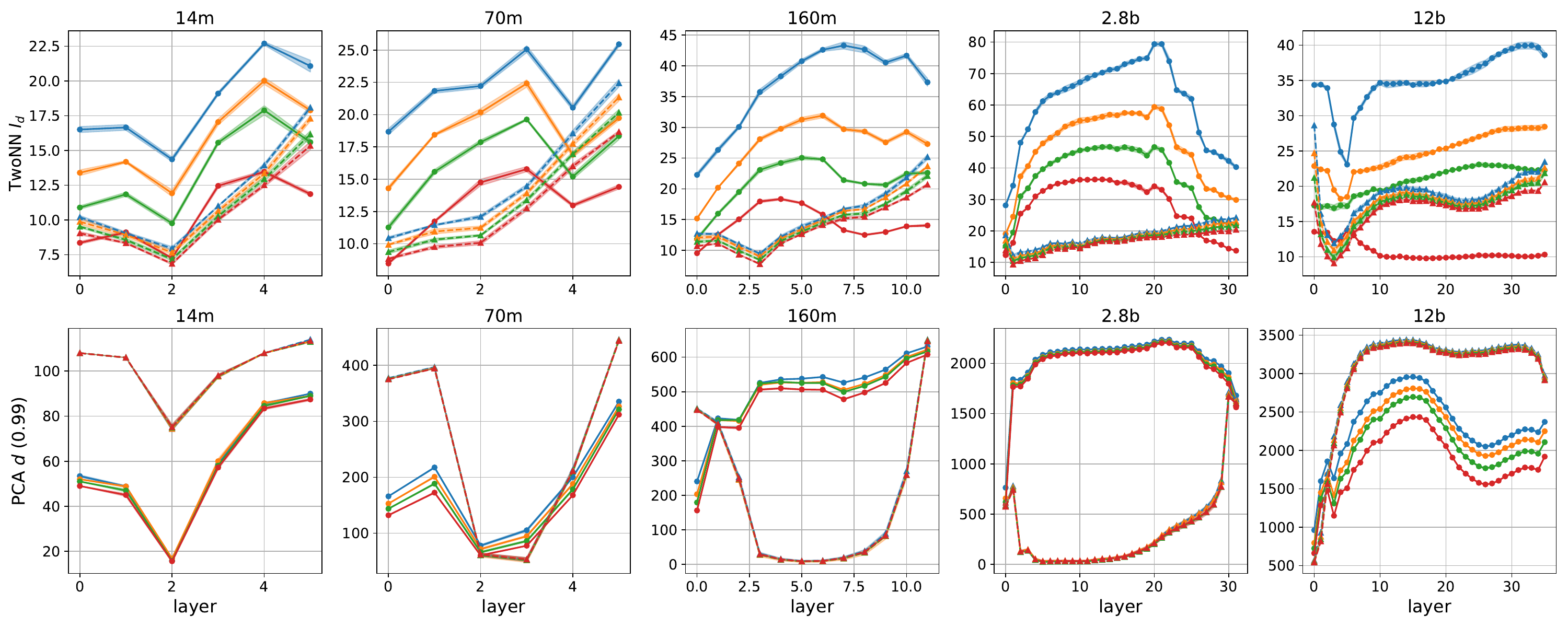}
    \caption{\textbf{Dimensionality over layers.} TwoNN nonlinear $I_d$ (top) and PCA linear $d$ (bottom) over layers are shown for all sizes (left to right). Each color corresponds to a coupling length $k\in 1\cdots 4$. solid curves denote coherent sequences, and dotted curves denote shuffled sequences. For all models, lower $k$ results in higher $I_d$ and $d$ for both normal and shuffled settings. For all models, shuffling results in lower $I_d$ but higher $d$. Curves are averaged over 5 random seeds, shown with $\pm1$ SD.}
    \label{fig:all_results}
\end{figure*}

\begin{figure*}[!htp]
    \centering
    \includegraphics[width=0.6\linewidth]{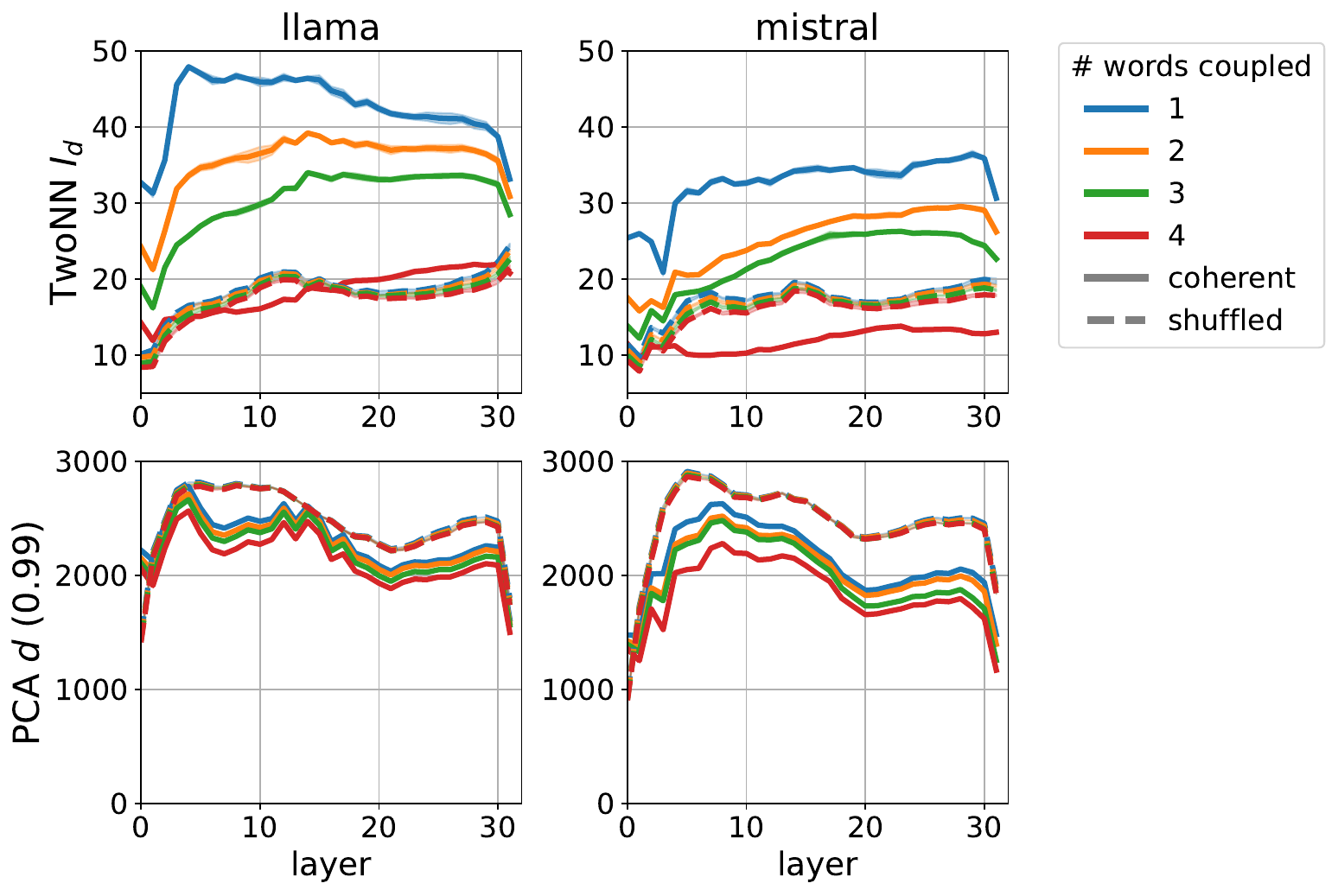}
    \caption{\textbf{Dimensionality over layers for Llama-3-8B and Mistral-7B.} TwoNN nonlinear $I_d$ (top) and PCA linear $d$ (bottom) over layers are shown for all sizes (left to right). Each color corresponds to a coupling length $k\in 1\cdots 4$. solid curves denote coherent sequences, and dotted curves denote shuffled sequences. For all models, lower $k$ results in higher $I_d$ and $d$ for both normal and shuffled settings. For all models, shuffling results in lower $I_d$ but higher $d$. Curves are averaged over 5 random seeds, shown with $\pm1$ SD. \textbf{Results mirror those of the Pythia models.}}
    \label{fig:other_models_results}
\end{figure*}

\begin{figure*}
    \centering
    \includegraphics[width=0.65\linewidth]{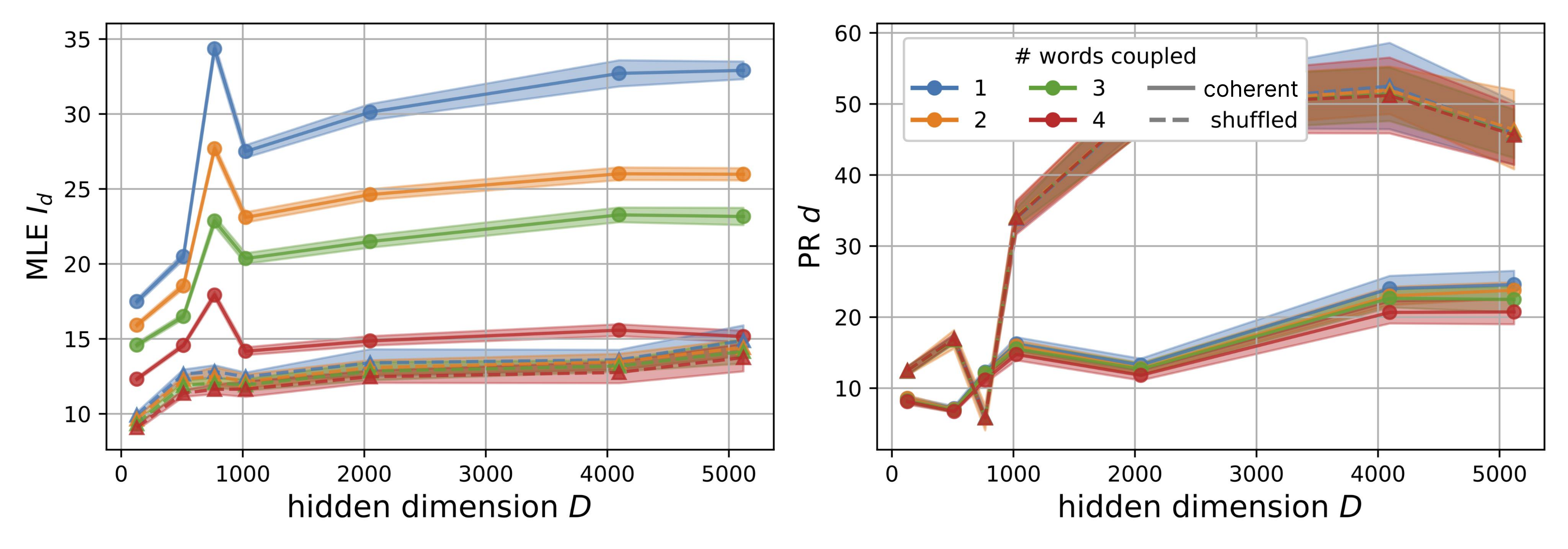}
    \caption{\textbf{Mean dimensionality over model size (other metrics).} Mean nonlinear $I_d$ computed with MLE (left) and linear $d$ computed with PR (right) over layers is shown for increasing LM hidden dimension $D$. MLE $I_d$ does not depend on extrinsic dimension $D$ (flat lines). PR $d$ produces nonsensical values, higher than the nonlinear $I_d$. Curves are averaged over 5 random seeds, shown with $\pm$ 1 SD.}
    \label{fig:pr_mle}
\end{figure*}

\begin{figure*}
    \centering
    \includegraphics[width=\linewidth]{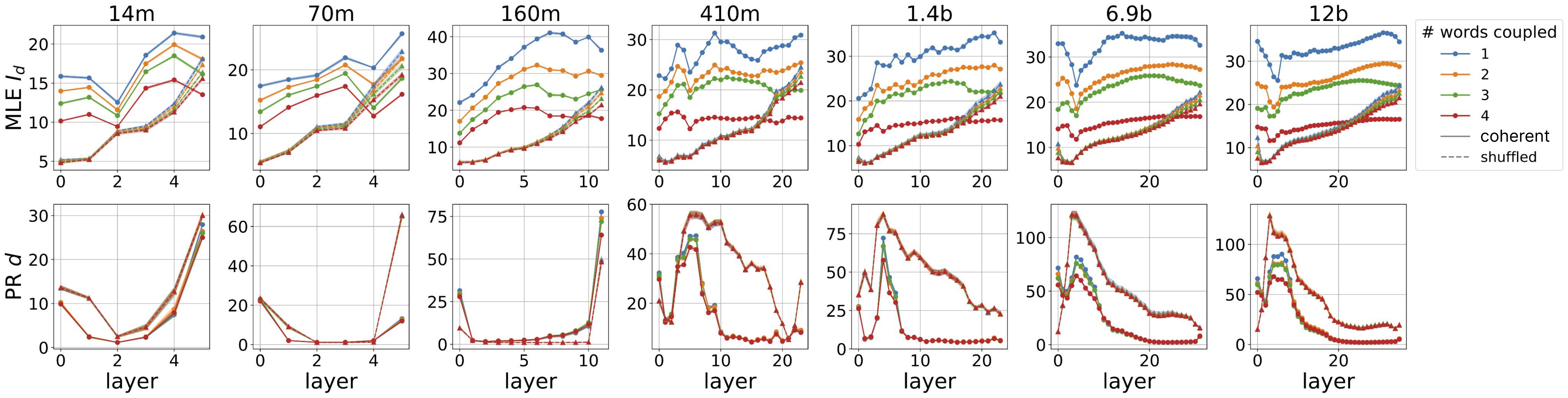}
    \caption{\textbf{Other dimensionality metrics over layers.} MLE nonlinear $I_d$ (top) and PR linear $d$ (bottom) over layers are shown for all model sizes (left to right). Each color corresponds to a coupling length $k\in 1\cdots 4$. solid curves denote coherent sequences, and dotted curves denote shuffled sequences. For all models, lower $k$ results in higher $I_d$ for both normal and shuffled settings. For all models, shuffling results in lower $I_d$. The PR-$d$ produced nonsensical results, with linear dimensionality higher than nonlinear dimensionality. Curves are averaged over 5 random seeds, shown with $\pm1$ SD.}
    \label{fig:mle_pr_over_layers}
\end{figure*}

\begin{figure}
    \centering
    \includegraphics[width=\linewidth]{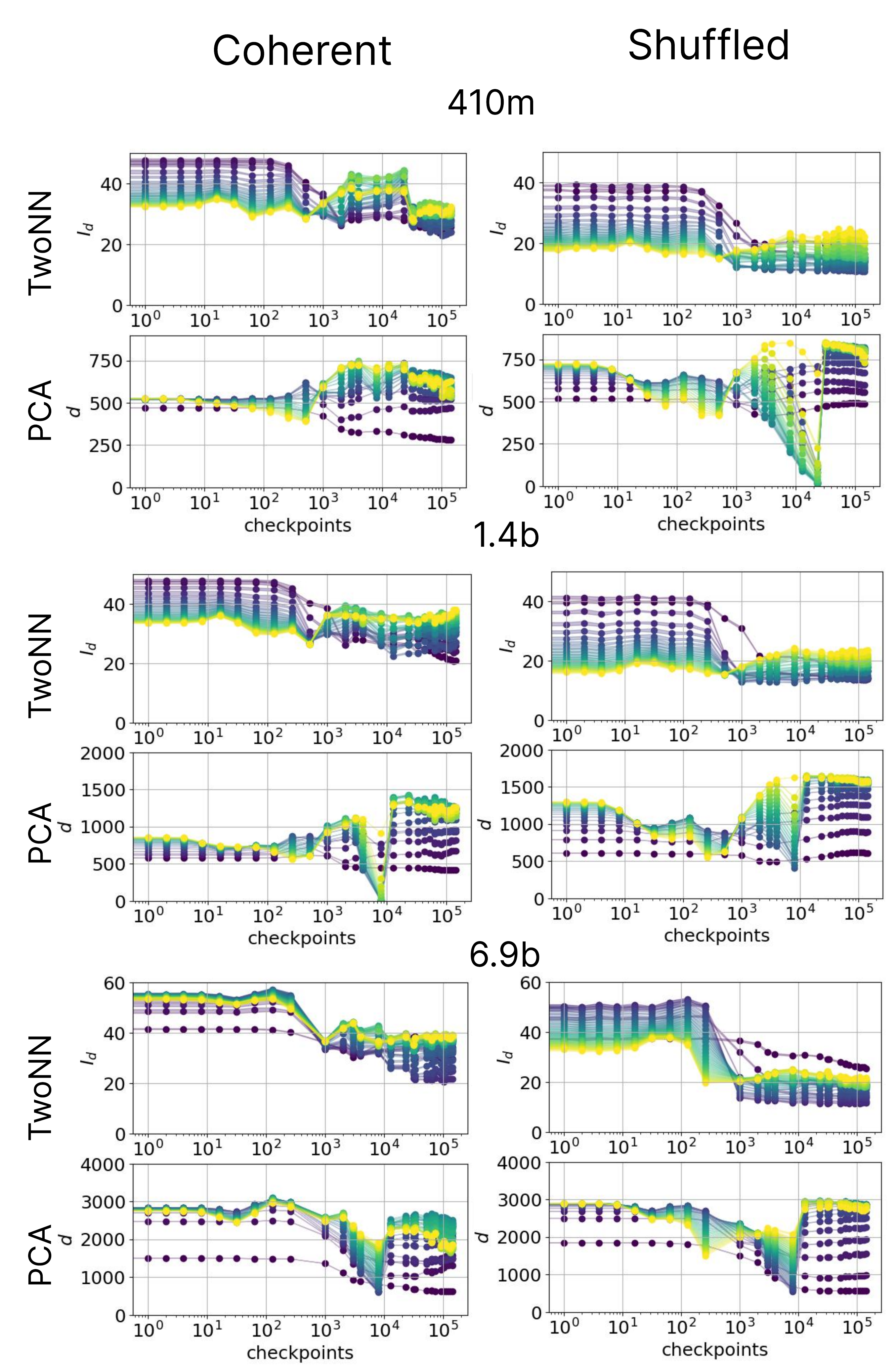}
    \caption{\textbf{Layerwise feature complexity evolution over time, additional results.} Nonlinear $I_d$ (top) and linear $d$ (bottom) over training is shown for coherent (left) and shuffled (right) text, for the $1$-coupled setting. Each curve is one layer of the LM (yellow is later, purple is earlier). All settings in [TwoNN, PCA]$\times$[coherent, shuffled] exhibit a phase transition in representational dimensionality at around checkpoint $10^3$, which corresponds to the sharp increase in task performance. In the nonlinear case (top row), the difference between layers' $I_d$ is \emph{low} at the end of training for shuffled text, and \emph{high} for coherent text. This suggests LM learns to perform meaningful and specialized processing over layers. The difference between layers' $d$ (bottom row) at the end of training is, conversely, \emph{high} for shuffled and \emph{lower} for coherent text. This is consistent with our interpretation of $d$ as capturing implied dataset size.}
    \label{fig:id_change_appendix}
\end{figure}

Here, we present extended results on the length-$17$ (longest) controlled grammar, for 
\begin{enumerate}
    \item All Pythia sizes, see \Cref{fig:all_results}, and Llama-3-8B and Mistral-7B, see \Cref{fig:other_models_results}.
    \item Other dimensionality measures MLE \citep{Levina_Bickel_2004} and Participation Ratio (PR) \citep{Gao2017ATO}, shown for Pythia models, see \Cref{fig:pr_mle,fig:mle_pr_over_layers}. Results are highly consistent between TwoNN and MLE, but PR produced nonsensical values $d_{PR} < I_d$ that violated the theoretical relation $I_d \leq d \leq D$.
    \item Linear fits $I_d \sim D$, $d\sim D$ for settings $k \in \{1\cdots 4\} \times \{\text{coherent}, \text{shuffled}\}$, see \Cref{tab:linear_fitting_D_PCA_twonn_prompts}.
    \item Training dynamics of feature complexity (TwoNN, PCA) for three Pythia sizes 410m, 1.4b, and 6.9b, on both coherent $k=1$ and shuffled text, see \Cref{fig:id_change_appendix}.
\end{enumerate}

\begin{table*}[!htbp]
\centering
\begin{tabular}{*8c}
\toprule
\textbf{Mode} & \textbf{$\mathbf{k}$-coupling} &  \multicolumn{3}{c}{\textbf{PCA} $d$} & \multicolumn{3}{c}{\textbf{TwoNN} $I_d$}\\
\midrule
{}   & {}   & $\boldsymbol{\alpha}$ & \textbf{R} & \textbf{p-value} & $\boldsymbol{\alpha}$& \textbf{R} & \textbf{p-value} \\
coherent & 1 & 0.4598 & 0.9956 & $2\times10^{-6}$ & 0.0023 & 0.6341 &0.1261\\
coherent & 2 & 0.4268& 0.9954 & $3\times10^{-6}$ & 0.0011& 0.5580 &0.1930 \\
coherent & 3& 0.4014 & 0.9943 & $5\times10^{-6}$ & 0.0009& 0.6616 &0.1056 \\
coherent & 4 & 0.3569 & 0.9924 & $1\times10^{-5}$ & -0.0003 & -0.3523 &0.4383 \\
shuffled & 1 & 0.6239& 0.9919 & $1.1\times10^{-5}$ & 0.0011 & 0.8488& 0.0157\\
shuffled & 2 & 0.6193 &0.9917  & $1.2\times10^{-5}$ & 0.0010 & 0.8487 &0.0157\\
shuffled & 3 & 0.6153 & 0.9916 & $1.2\times10^{-5}$ & 0.0010& 0.8586 & 0.0134\\
shuffled & 4 & 0.6114 &0.9916 & $1.2\times10^{-5}$ & 0.0009 & 0.8559 & 0.0140\\
\bottomrule
\end{tabular}
\caption{\textbf{Linear regression of average layerwise dimensionality to hidden dimension, $D$}. For Pythia models, for each setting (Mode, $k$-coupling) and dimensionality measure (PCA, TwoNN), the linear effect size $\alpha$ along with $R$-value and $p$-value are reported. PCA linear dimension shows a consistent strong linear relationship with large effect size $\alpha$ to hidden dimension $D$ ($p<0.001$) for all settings in $k=\{1\cdots 4\} \times [\text{coherent, shuffled}]$. TwoNN intrinsic dimension does not scale linearly as $D$ in all settings, showing a non-significant relationship for coherent text and a significant one for shuffled text. For all TwoNN settings, the effect size $\alpha$ is near-zero, showing that nonlinear $I_d$ is robust to changes in hidden dimension $D$.}
\label{tab:linear_fitting_D_PCA_twonn_prompts}
\end{table*}

\newpage
\section{Additional Results: The Pile}
\label{app:pile}
\setcounter{figure}{0}    
\setcounter{table}{0} 

\begin{figure}[h!]
    \centering
    \includegraphics[width=\linewidth]{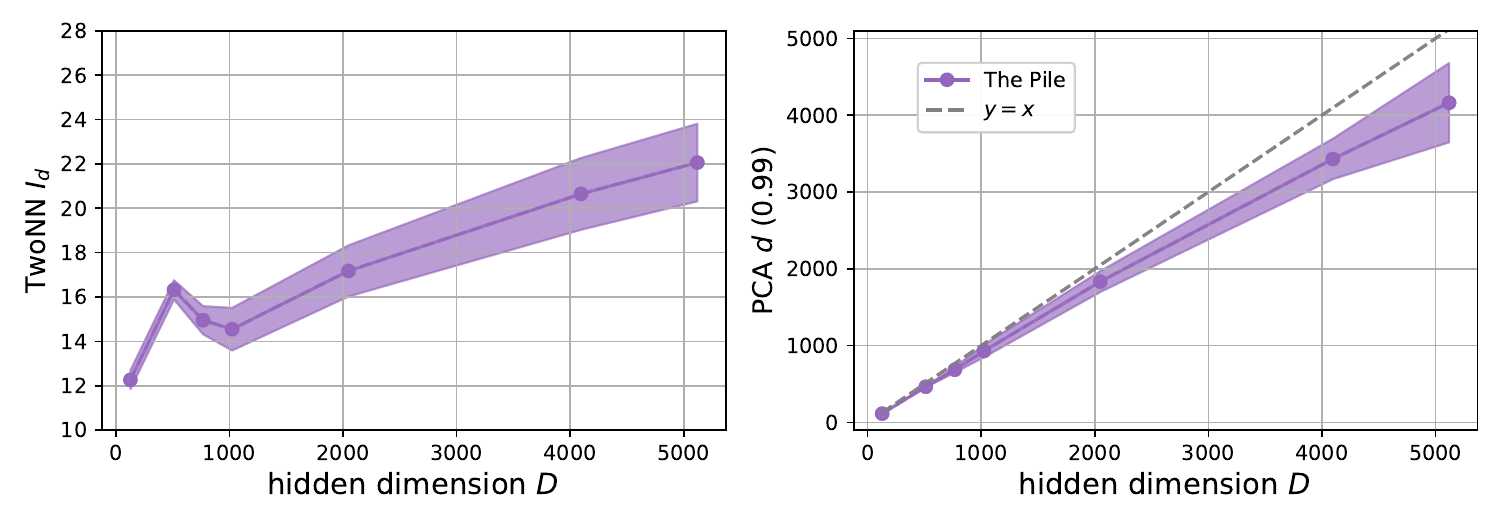}
    \caption{\textbf{Mean dimensionality on the Pile over model size.} Mean nonlinear $I_d$ computed with TwoNN (left) and linear $d$ computed with PCA (right) over layers is shown for increasing LM hidden dimension $D$. TwoNN $I_d$ grows very slowly with extrinsic dimension $D$, while PCA $d$ grows to be nearly one-to-one with $D$. Curves are averaged over 5 random data splits, shown with $\pm$ 1 SD.}
    \label{fig:pile-id-model-size-app}
\end{figure}

Similar to for the controlled grammar (\Cref{sec:results-model-size}), representational $I_d$ and $d$ scale differently with model ambient dimension $D$, where we found for the controlled grammar that $d\propto D$ while $I_d$ is stable in $D$. Here, \Cref{fig:pile-id-model-size-app} and \Cref{tab:linear_fitting_D_PCA_twonn_pile} similarly show that $d\propto D$ with a highly significant linear fit; a large effect $\alpha=0.81$ indicates the LM fills the ambient space such that $d \approx D$. For The Pile, $I_d \propto D$ ($R = 0.95$, $p < 0.001$) as well, but the tiny effect $\alpha$=$0.002$ shows $I_d$ is stable as $D$ grows, see \Cref{fig:pile-id-model-size-app} (left).

\begin{table*}[!htbp]
\centering
\begin{tabular}{*6c}
\toprule
\multicolumn{3}{c}{\textbf{PCA} $d$} & \multicolumn{3}{c}{\textbf{TwoNN} $I_d$}\\
\midrule
$\boldsymbol{\alpha}$ & \textbf{R} & \textbf{p-value} & $\boldsymbol{\alpha}$& \textbf{R} & \textbf{p-value} \\
0.8119 & 0.9993 & $2.39\times10^{-8}$ & 0.00173 & 0.9537 &$8.64\times10^{-4}$\\
\bottomrule
\end{tabular}
\caption{\textbf{Linear regression of Pythia's average layerwise dimensionality on The Pile to hidden dimension, $D$}. For dimensionality measures (PCA, TwoNN columns), the linear effect size $\alpha$ along with $R$-value and $p$-value are reported. PCA linear dimension shows a statistically significant linear relationship to $D$, with large effect size $\alpha = 0.81$. TwoNN intrinsic dimension also shows a slightly weaker, but still highly significant, linear relationship to $D$. But, the effect size $\alpha$ is near-zero, showing that nonlinear $I_d$ is robust to changes in hidden dimension $D$.}
\label{tab:linear_fitting_D_PCA_twonn_pile}
\end{table*}

\begin{figure*}[h!]
    \centering
    \includegraphics[width=1\linewidth]{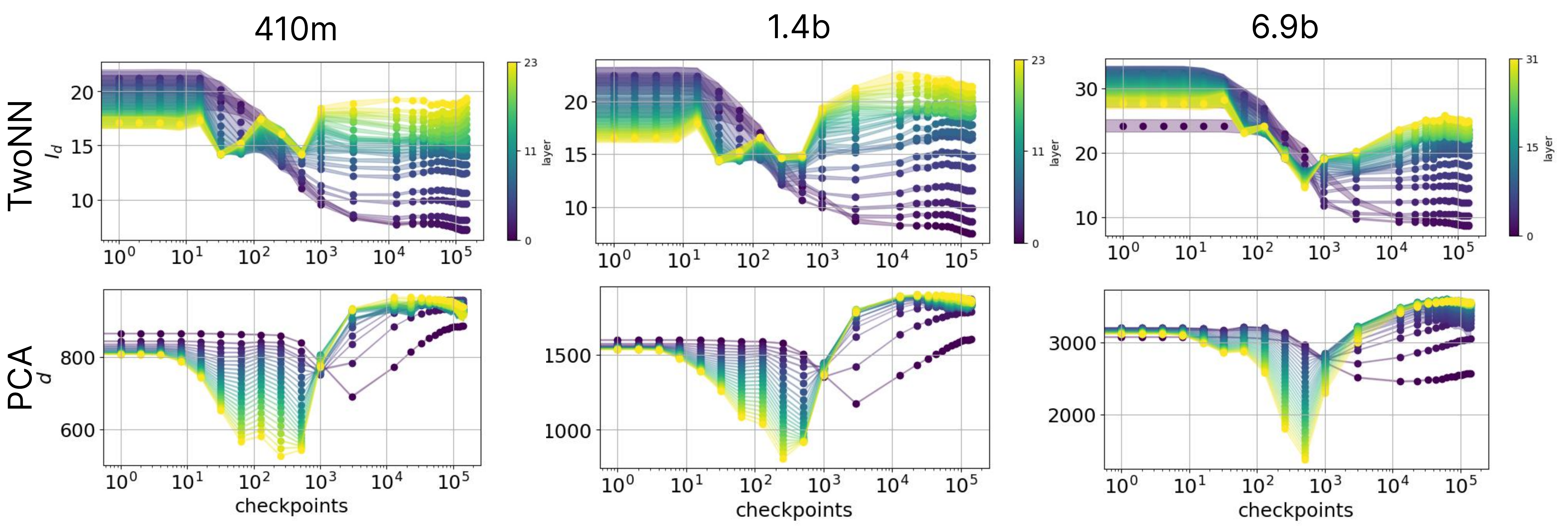}
    \caption{\textbf{$I_d$ phase transition in The Pile.} Nonlinear $I_d$ (top) and linear $d$ (bottom) over training is shown for model sizes 410m, 1.4b, and 6.9b (left to right), for The Pile. Each curve is one layer of the LM (yellow is later, purple is earlier). Representations of The Pile exhibit a phase transition in both $I_d$ and $d$ at slightly before checkpoint $10^3$, where $t=10^3$ corresponds to a dip and redistribution of layerwise dimensionality, and also a sharp increase in task performance in \Cref{fig:perf-id}.}
    \label{fig:pile_id_change_appendix}
\end{figure*}

\section{Additional Results: Correlation with Kolmogorov Complexity}
\setcounter{figure}{0}    
\setcounter{table}{0} 
\label{app:kc}

\subsection{Correlation between Kolmogorov Complexity and Feature Complexity is Robust to Sequence Length}
In \Cref{tab:gzip} we showed that, for each model, and on a single dataset ($k=1$, $l=17$), linear effective $d$ highly correlates to the estimated superficial complexity (KC) using \texttt{gzip}. \Cref{tab:gzip_extended} shows that this trend is robust to both model family, model size, and sequence length; average layerwise $d$ is almost perfectly monotonic in the KC of the dataset, seen by high Spearman correlation. In contrast, for none of the sequence lengths is average layerwise $I_d$ monotonic in KC, except for the smallest Pythia model (14m).

\subsection{Kolmogorov Complexity vs.~Average-Layer Feature Complexity Across Datasets}
\begin{figure*}
    \centering
    \includegraphics[width=0.9\linewidth]{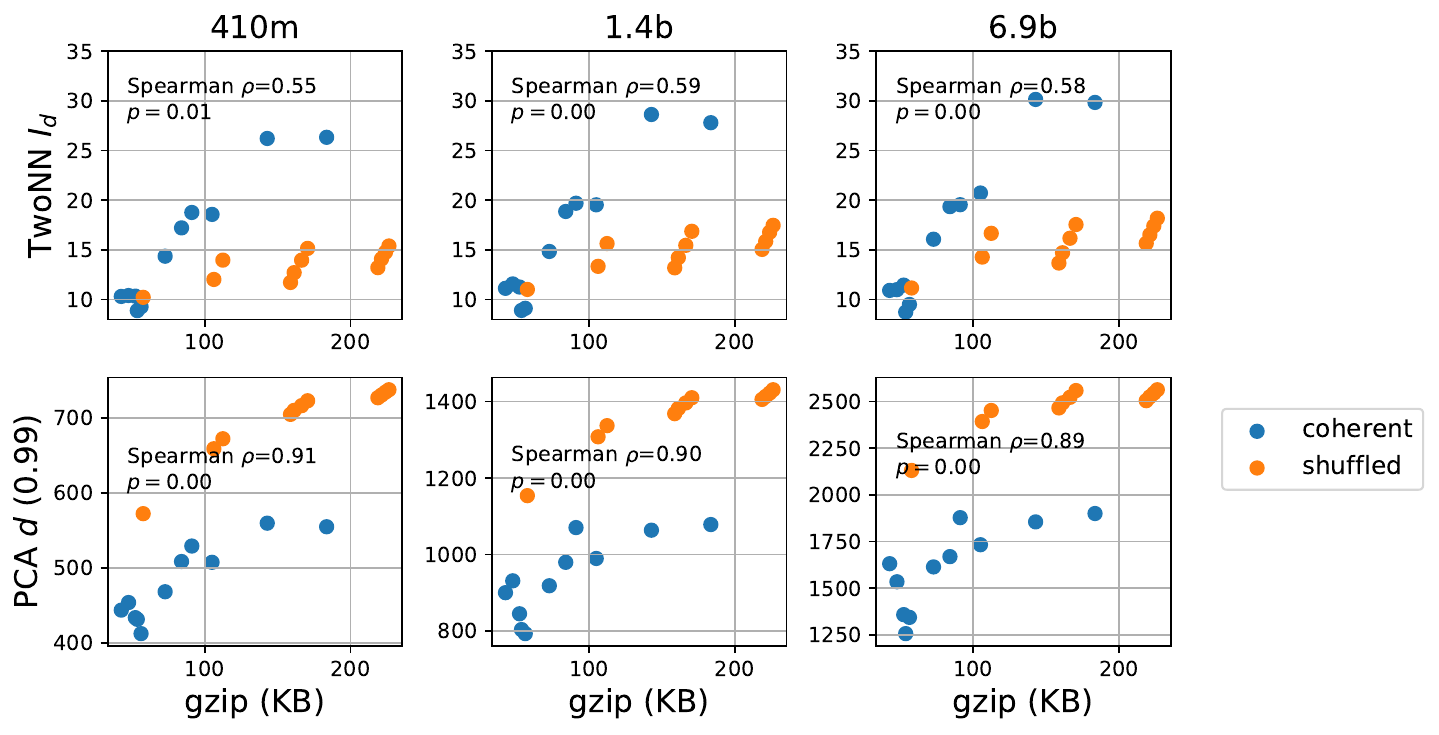}
    \caption{\textbf{Average layerwise dimensionality vs.~Estimated Kolmogorov Complexity (gzip)} for Pythia 410m, 1.4b, and 6.9b, aggregated for all grammars. For all models, PCA $d$ highly correlates to \texttt{gzip} (estimated KC), with Spearman $\rho \geq 0.9$** for all models. TwoNN $I_d$ correlates more weakly, $\rho \in [0.5, 0.6]$* for all models. Linear $d$ and nonlinear $I_d$ differentially encode shuffled data complexity (orange dots) compared to coherent data complexity (blue dots); where shuffled data display higher $d$ and lower $I_d$. (**) Significant at $\alpha=0.001$, (*) $\alpha=0.01$.}
    \label{fig:kc_all_settings}
\end{figure*}

\Cref{fig:kc_all_settings} shows the global correlation between feature complexity ($I_d$ and $d$) and KC, estimated with \texttt{gzip}. While both nonlinear (top row) and linear (bottom row) dimensionality are positively Spearman-correlated to \texttt{gzip}, there are clear differences:
\begin{enumerate}
    \item Linear $d$ increases in the shuffled setting from the coherent setting; nonlinear $I_d$ decreases.
    \item Linear $d$ is very highly correlated to the estimated KC, $\rho \approx 0.9$ in all cases, while nonlinear $d$ is more weakly correlated, $\rho \in [0.5, 0.6]$.
\end{enumerate}

These observations support the hypothesis that linear effective $d$ encodes KC, while the intrinsic dimension $I_d$ encodes sequence-level semantic complexity.

\begin{table*}
\centering 
\caption{\textbf{Spearman correlations between dimensionality and estimated Kolmogorov complexity, varying sequence length}. The Spearman correlation $\rho$ between the 
\texttt{gzip}ped dataset size (KB) and representational dimensionality (rows), averaged over layers, is shown for all tested \textbf{Pythia} model sizes (model name omitted for readability). Values marked with a * are significant with a p-value threshold of 0.05. Values marked with \textdagger \space are significant with a p-value threshold of 0.1. Across models, average-layer $I_d$ is not correlated to the estimated KC, or formal compositionality, of datasets. Average-layer \textbf{linear $d$} is consistently highly positively correlated to the estimated KC. Length $l=5$ is grayed out as, due to the sequence length being too short, it was not possible to varying the coupling factor $k$; here, the only comparison is between coherent and shuffled ($n=2$). \\ }
\begin{tabular}{ccccccc}
\hline
\multirow{2}{*}{} & \multirow{2}{*}{} & \multicolumn{5}{c}{sequence length (words)} \\
\cline{3-7}
 & & \textcolor{gray}{5} & 8 & 11 & 15 & 17 \\ 
\hline
\multirow{2}{*}{14m} 
 & $I_d$ & \textcolor{gray}{1.00} & 0.89 & \textbf{0.87}\(^*\) & \textbf{0.87}\(^*\) & -0.10 \\ 
 & $d$ & \textcolor{gray}{1.00} & 0.89 & \textbf{0.87}\(^*\) & \textbf{0.87}\(^*\) & \textbf{0.81}\(^*\) \\ 
\cline{3-7}
\multirow{2}{*}{70m} 
 & $I_d$ & \textcolor{gray}{1.00} & 0.40 & 0.43 & 0.26 & -0.10 \\ 
 & $d$ & \textcolor{gray}{1.00} & \textbf{1.00}\(^*\) & \textbf{1.00}\(^*\) & \textbf{0.98}\(^*\) & \textbf{0.98}\(^*\) \\ 
\cline{3-7}
\multirow{2}{*}{160m} 
 & $I_d$ & \textcolor{gray}{-1.00} & 0.00 & 0.19 & 0.00 & -0.21 \\ 
 & $d$ & \textcolor{gray}{-1.00} & -0.60 & -0.52 & -0.62 & -0.62 \\ 
\cline{3-7}
\multirow{2}{*}{410m} 
 & $I_d$ & \textcolor{gray}{-1.00} & 0.40 & 0.40 & 0.26 & 0.14 \\ 
 & $d$ & \textcolor{gray}{1.00} & \textbf{1.00}\(^*\) & \textbf{0.98}\(^*\) & \textbf{1.00}\(^*\) & \textbf{1.00}\(^*\) \\ 
\cline{3-7}
\multirow{2}{*}{1.4b} 
 & $I_d$ & \textcolor{gray}{-1.00} & 0.40 & 0.40 & 0.43 & 0.14 \\ 
 & $d$ & \textcolor{gray}{1.00} & \textbf{1.00}\(^*\) & \textbf{0.98}\(^*\) & \textbf{1.00}\(^*\) & \textbf{1.00}\(^*\) \\ 
\cline{3-7}
\multirow{2}{*}{6.9b} 
 & $I_d$ & \textcolor{gray}{1.00} & 0.40 & 0.40 & 0.36 & 0.48 \\ 
 & $d$ & \textcolor{gray}{1.00} & \textbf{1.00}\(^*\) & \textbf{0.98}\(^*\) & \textbf{1.00}\(^*\) & \textbf{0.98}\(^*\) \\ 
\cline{3-7}
\multirow{2}{*}{12b} 
 & $I_d$ & \textcolor{gray}{1.00} & 0.40 & 0.40 & 0.43 & 0.00 \\ 
 & $d$ & \textcolor{gray}{1.00} & \textbf{1.00}\(^*\) & \textbf{0.98}\(^*\) & \textbf{1.00}\(^*\) & \textbf{1.00}\(^*\) \\ 
\cline{3-7}
\multirow{2}{*}{Llama-8b} 
 & $I_d$ & \textcolor{gray}{1.00} & 0.40 & 0.19 & 0.00 & -0.02 \\ 
 & $d$ & \textcolor{gray}{1.00} & \textbf{1.00}\(^*\) & \textbf{0.98}\(^*\) & \textbf{1.00}\(^*\) & \textbf{0.93}\(^*\) \\ 
\cline{3-7}
\multirow{2}{*}{Mistral-7b} 
 & $I_d$ & \textcolor{gray}{1.00} & 0.40 & 0.40 & 0.00 & 0.29 \\ 
 & $d$ & \textcolor{gray}{1.00} & \textbf{1.00}\(^*\) & \textbf{0.98}\(^*\) & \textbf{1.00}\(^*\) & \textbf{0.90}\(^*\) \\ 
\hline
\end{tabular}

\label{tab:gzip_extended}
\end{table*}

\subsection{Per-layer Correlation with Kolmogorov Complexity}

\begin{figure*}[h!]
    \centering
    \includegraphics[width=\linewidth]{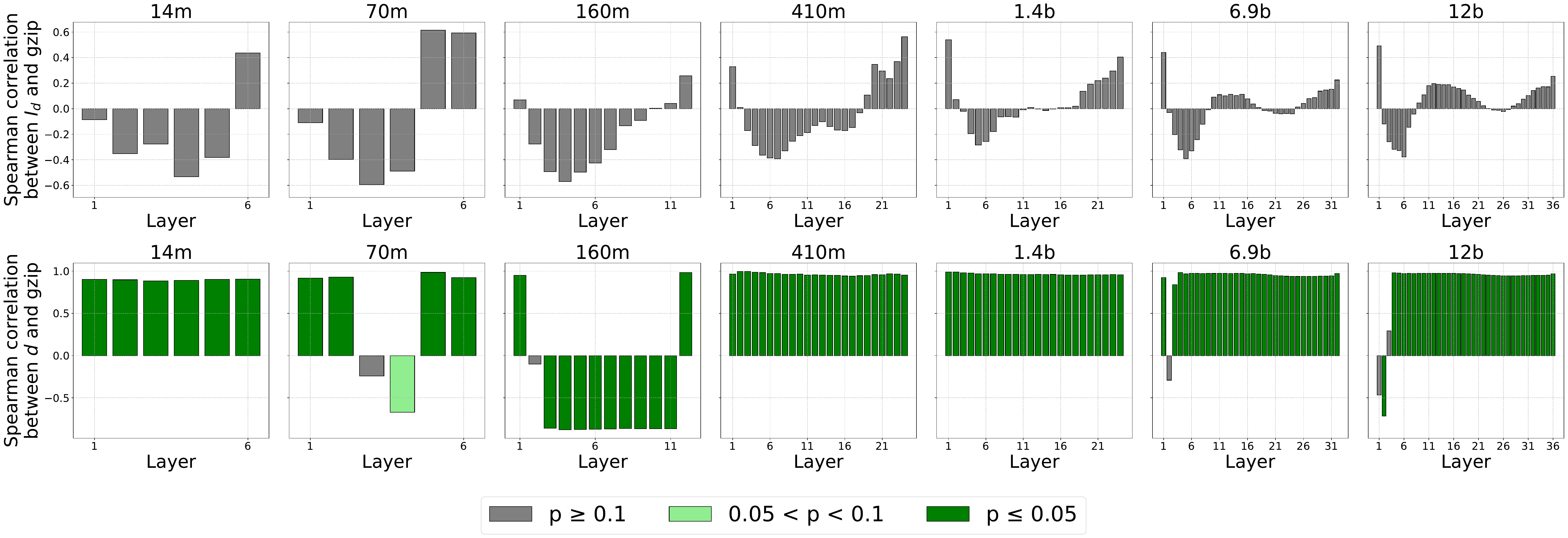}
    
    \caption{\textbf{Spearman correlations between per-layer dimensionality and estimated Kolmogorov complexity, Pythia models.}
The Spearman correlation between the gzipped dataset size (KB) and representational dimensionality per layer, is shown for all tested model sizes for the longest sequence length ($l=17$). Generally across models, per-layer $I_d$ is not correlated to the estimated KC, or formal compositionality, of datasets. Per-layer linear $d$ is consistently highly positively correlated to the estimated Kolmogorov complexity, except one outlier (160m).}
    \label{fig:layer-correlations}
\end{figure*}

\begin{figure}
    \centering
    \includegraphics[width=\linewidth]{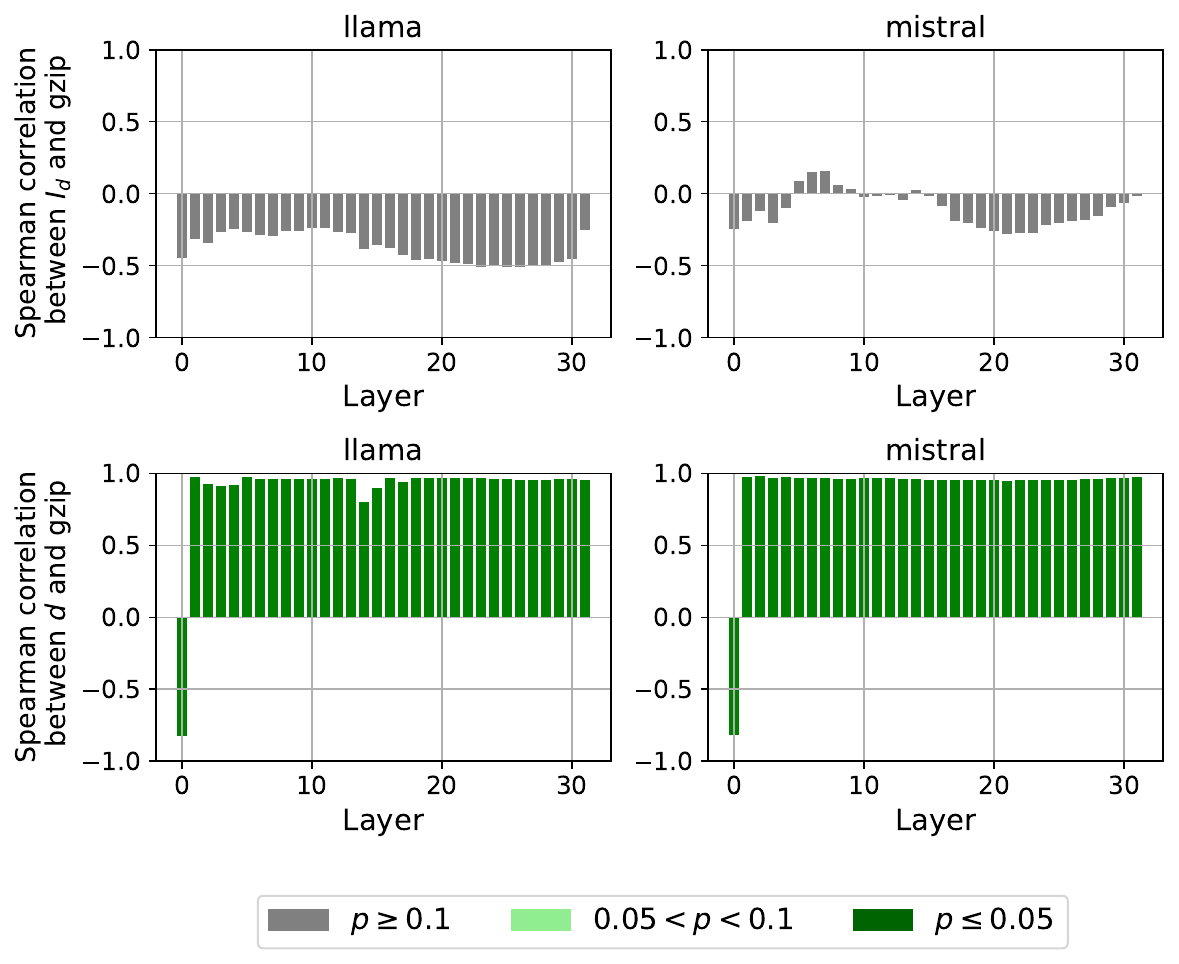}
    \caption{\textbf{Spearman correlations between per-layer dimensionality and estimated Kolmogorov complexity, Llama-3-8B and Mistral-7B.}
The Spearman correlation between the gzipped dataset size (KB) and representational dimensionality per layer, is shown for Llama (left) and Mistral (right). Consistently across models, mirroring trends for Pythia, per-layer $I_d$ is not correlated to the estimated KC, or formal compositionality, of datasets. Per-layer linear $d$ is consistently highly positively correlated to the estimated KC.}
\label{fig:other_models_layer_correlations}
\end{figure}

\begin{figure*}
    \centering
    \includegraphics[width=\linewidth]{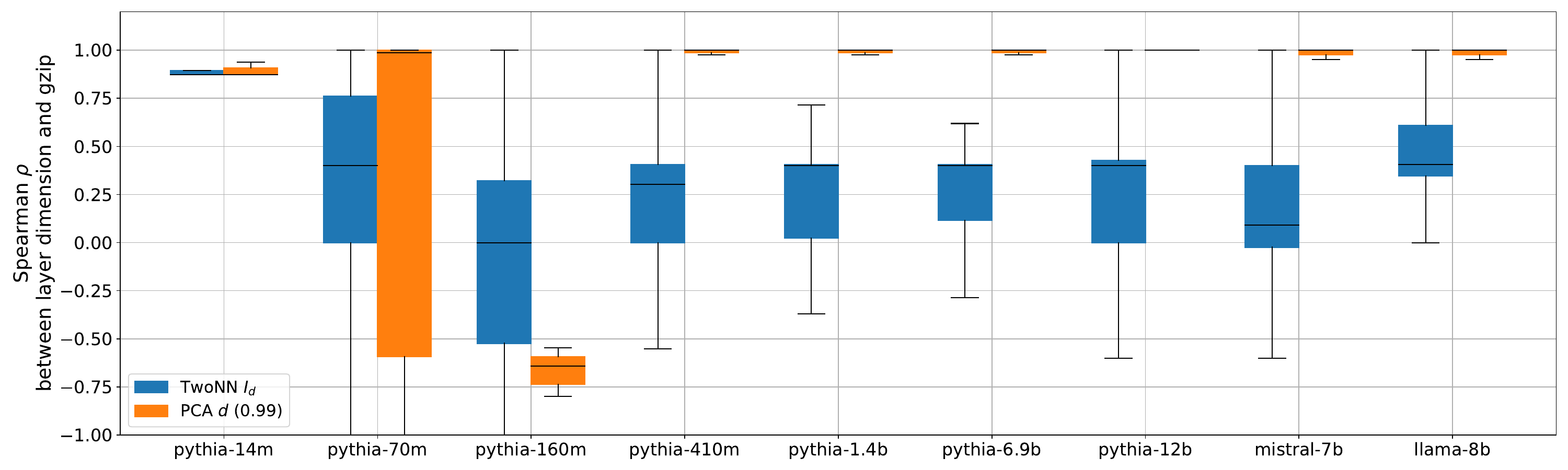}
    \caption{\textbf{Per-layer Spearman $\rho$ between feature complexity and KC for all models, across all tested datasets.} The layerwise Spearman $\rho$ between KC, measured with \texttt{gzip}, and feature complexity, measured with TwoNN $I_d$ (blue) and PCA $d$ (orange), is shown for each model. Each datapoint in each distribution corresponds to one (model, dataset, layer) triple. Generally across models, except for the outlier Pythia-160m, the layerwise correlation between $I_d$ and KC is low, while the correlation to $d$ is high and close to $1.0$ for the vast majority of layers, datasets, and models (orange distributions near $1.0$). This shows that, with high generality across models and datasets, the vast majority of layers encode KC in linear effective $d$, not in the intrinsic dimension $I_d$. Trends are especially robust after a certain model size ($\geq$410m).}
    \label{fig:kc_overall_summary}
\end{figure*}

\begin{figure*}[h!]
    \centering
    \begin{subfigure}[t]{0.45\linewidth}
        \centering
        \includegraphics[width=\textwidth]{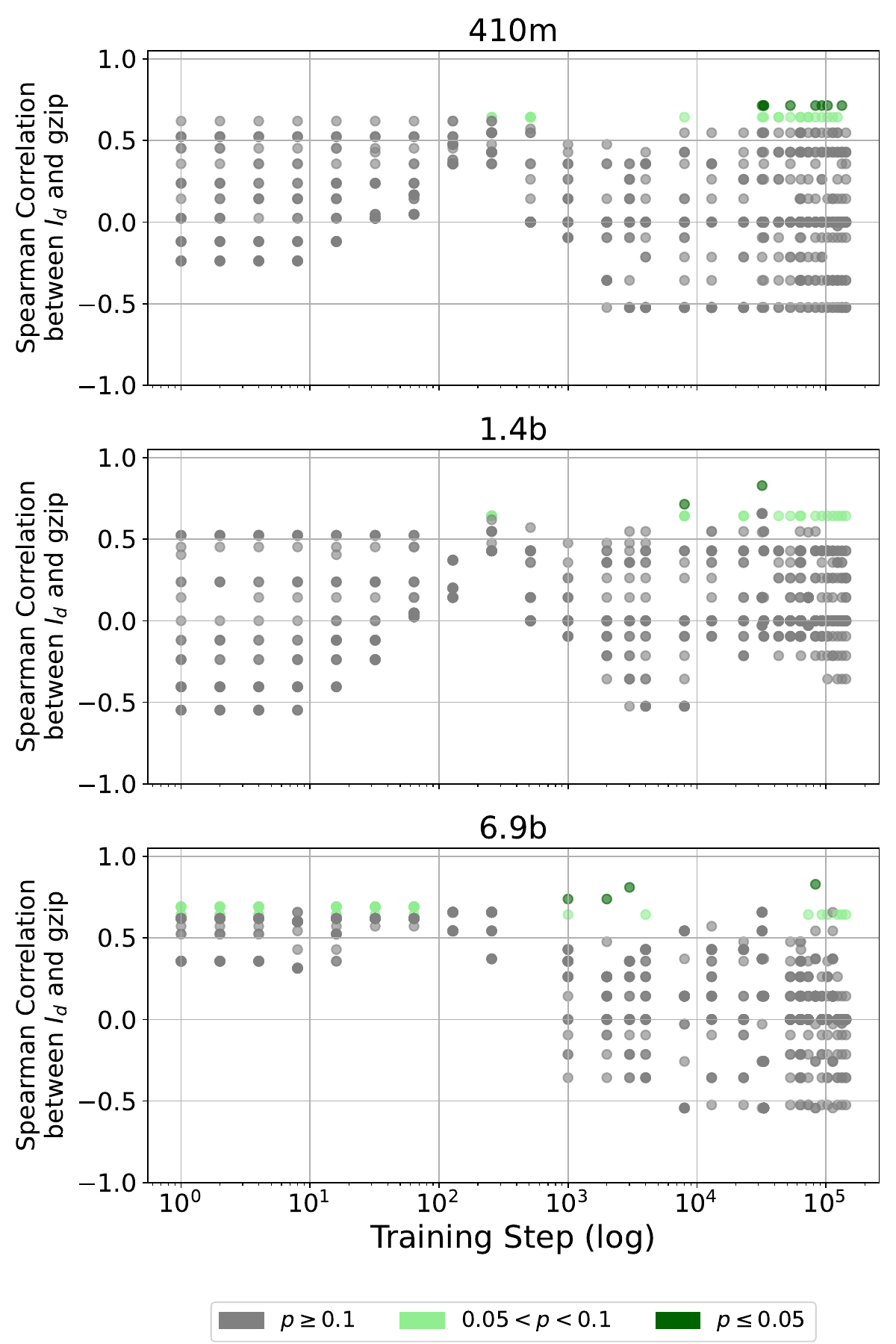}
        \caption{TwoNN $I_d$}
        \label{fig:twonn-evol-kc}
    \end{subfigure}
    \hfill
    \begin{subfigure}[t]{0.45\linewidth}
        \centering
        \includegraphics[width=\textwidth]{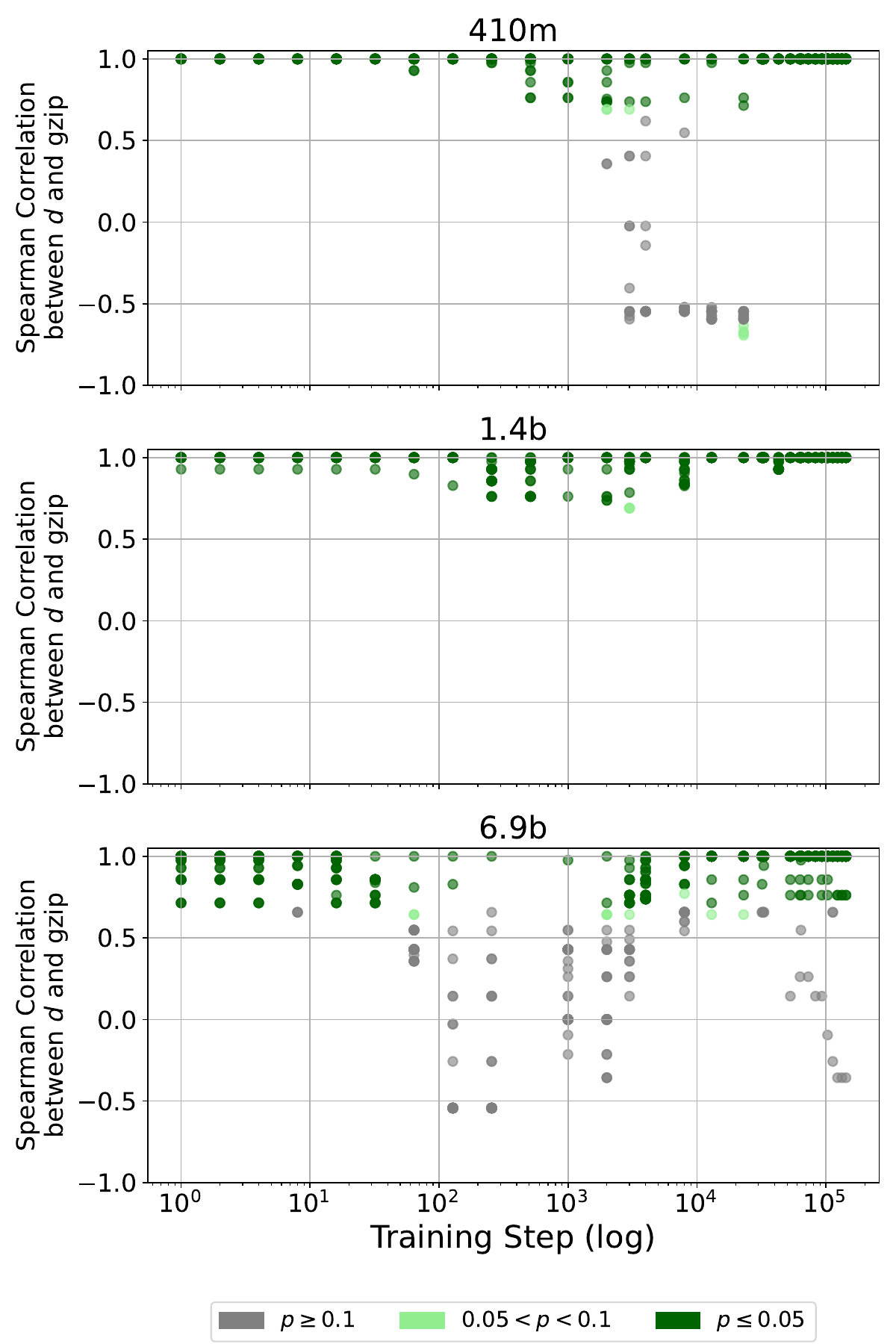}
        \caption{PCA $d$}
        \label{fig:pca-evol-kc}
    \end{subfigure}
    \caption{\textbf{Spearman correlation of layerwise dimensionality and Kolmogorov complexity over training.} The Spearman correlations between $I_d$ (left) and \texttt{gzip}, and $d$ (right) and \texttt{gzip} are plotted for three models: 410m, 1.4b, and 6.9b (top to bottom) over training time (x-axis). Correlations are computed across the controlled corpora. Each vertical set of points denotes the layer distribution of Spearman correlations at a single timestep; each point is one layer's Spearman correlation, colored green if statistically significant and gray otherwise.}
    \label{fig:time-evolution-kc}
\end{figure*}

\paragraph{Linear effective $d$ encodes KC robustly across models and datasets} \Cref{fig:kc_overall_summary} shows, for each model, the Spearman correlation between layer dimension and KC (\texttt{gzip}). Orange boxplots correspond to $d$, and blue boxplots to the $I_d$. Each datapoint in a boxplot reports the correlation for one (model, layer, sequence length) combination; the only factor of variation in each correlation is the $k$-coupling factor and whether the dataset is shuffled. \textbf{With high generality across models and grammars, the linear effective $d$ is monotonic in KC}, seen by the vast majority of layers (orange distributions) close to $\rho=1.0$ (y-axis). Meanwhile, the $I_d$ does not consistently encode superficial complexity, seen by the blue distributions landing about $0.0$. 

\paragraph{Outliers} There was one significant outlier, 160m, in our analysis correlating layerwise dimensionality to \texttt{gzip} (KC), see \Cref{fig:layer-correlations,fig:kc_overall_summary} and \Cref{tab:gzip_extended}. While other models consistently demonstrate a positive Spearman correlation between $d$ and \texttt{gzip} across layers, 160m (and to a smaller extent, 70m) deviates from this pattern. The reason 160m displays a negative correlation is due to its behavior on shuffled corpora, see the third column in \Cref{fig:all_results}: for intermediate layers, PCA with a variance threshold of 0.99 yields fewer than 50 PCs. We found that this was due to the existence of so-called ``rogue dimensions" \citep{timkey-van-schijndel-2021-bark,machina-mercer-2024-anisotropy,rudman-etal-2023-outlier}, where very few dimensions have outsized norms. Outlier dimensions have been found, via mechanistic interpretability analyses, to serve as a ``sink" for uncertainty, and are associated to very frequent tokens in the training data \citep{puccetti-etal-2022-outlier}. See \citet{rudman-etal-2023-outlier} for exact activation profiles for the last-token embeddings in Pythia 70m and 160m. While increasing the variance threshold to $0.999$ reduced the effect of rogue dimensions on PCA dimensionality estimation, we decided to keep the threshold at $0.99$ for consistent comparison to other models. 

\paragraph{Coding of superficial complexity over training} The Spearman correlations between layerwise dimensionality ($I_d$ and $d$) and estimated Kolmogorov complexity using \texttt{gzip}, over training steps, are shown in \Cref{fig:time-evolution-kc} for 410m, 1.4b, and 6.9b. Each dot in the figure is a single layer's correlation to \texttt{gzip}; each vertical set of dots is the distribution of correlations over layers, at a single timestep of training. Several observations stand out:

\begin{enumerate}
    \item PCA encodes superficial complexity (seen by earlier dots close to $\rho=1.0$) as an inductive bias of the model architecture. The high correlation for most layers may be unlearned during intermediate checkpoints of model training, seen by the ``dip" in gray dots around steps $10^2\sim 10^3$, but is regained by the end of training for all model sizes. This indicates that encoding superficial complexity at the end of training is a \emph{learned behavior}.
    \item TwoNN $I_d$ does not statistically significantly correlate to \texttt{gzip} at any point during training, for virtually all layers.
    \item For $I_d$, the phase transition noted in \Cref{sec:phase_transition} is also present at slightly before $t=10^3$; this is seen by layerwise correlations in \Cref{fig:time-evolution-kc}a coalescing to around $\rho=0.5$, and then redistributing. The layers that best encode superficial complexity for TwoNN at the end of trianing correspond to model-initial and model-final layers, see \Cref{fig:layer-correlations} top.
\end{enumerate}


\vfill
\pagebreak 
\section{Additional Results: Varying Sequence Length}
\label{app:sequence_length}
\setcounter{figure}{0}    
\setcounter{table}{0} 

\begin{figure*}
    \includegraphics[width=\linewidth]{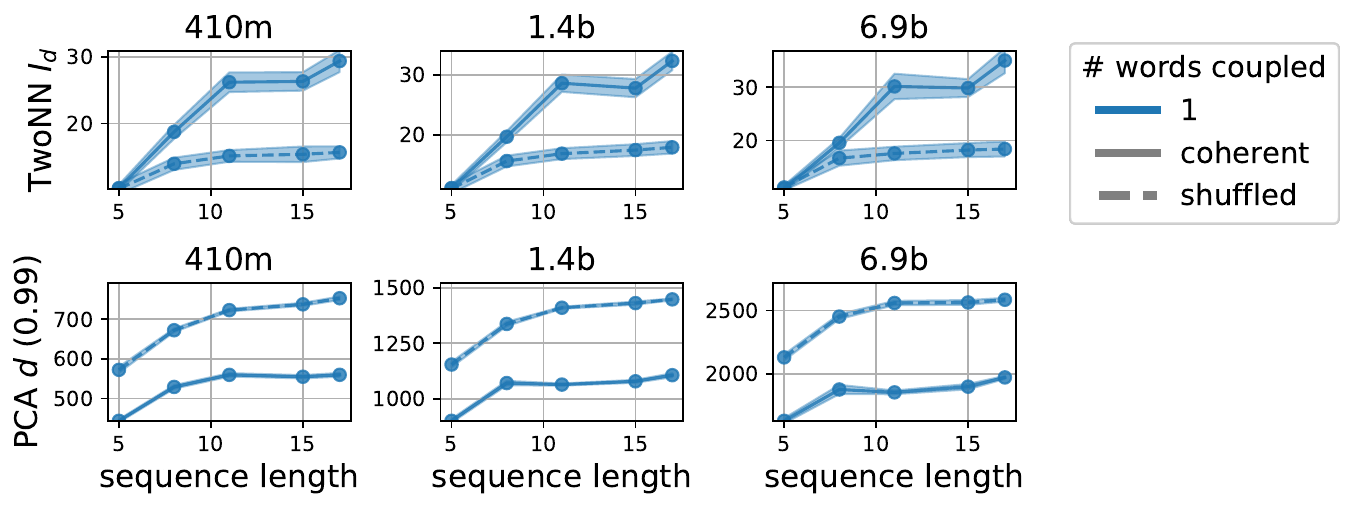}
    \caption{\textbf{Feature complexity increases over sequence length}. The mean $I_d$ and $d$ over layers (y-axis) is shown for increasing sequence lengths $\in \{5, 8, 11, 15, 17\}$ (x-axis) for Pythia models $\in \{$410m, 1.4b, 6.9b$\}$ (left to right), for the $k=1$, or the original dataset configuration. solid curves correspond to coherent, and dashed to shuffled, text. All curves are shown $\pm 1$SD over 5 random seeds. Y-axes are scaled to the minimum and maximum for each plot for readability. All curves increase from left to right, evidencing that both nonlinear and linear feature complexity increase with sequence length. Moreover, all curves \emph{saturate}, or plateau, around length=$11$, indicating this dependence is sublinear.}
    \label{fig:ramping}
\end{figure*}

\begin{figure*}
    \centering
    \includegraphics[width=0.5\textwidth]{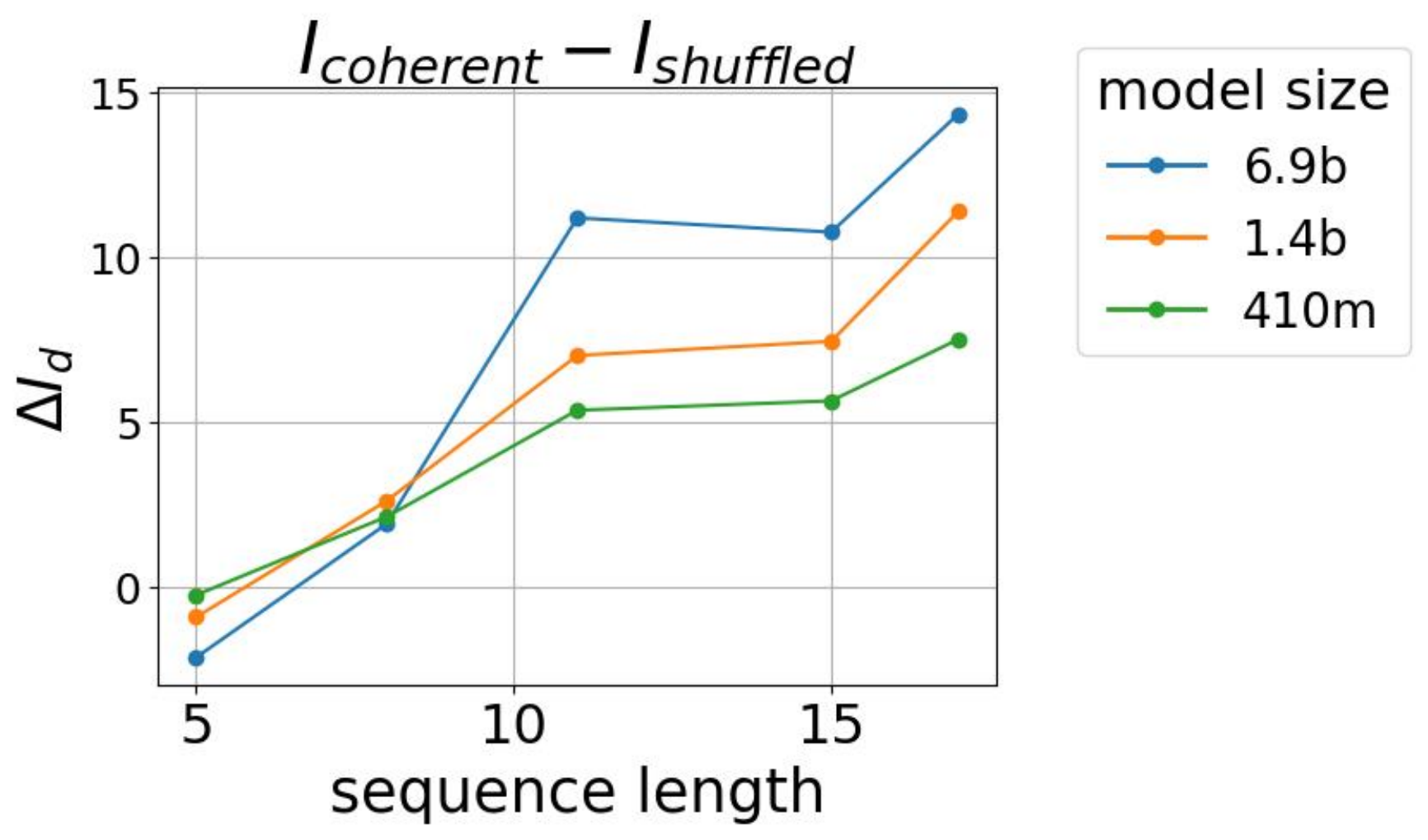}
    \caption{\textbf{Differential coding of semantic complexity increases with sequence length}. The $\Delta(I_d)$ between coherent and shuffled text (y-axis) is shown for Pythia models $\in \{$410m, 1.4b, 6.9b$\}$ (different curves), as a function of sentence length $\in \{5, 8, 11, 15, 17\}$, (x-axis). For all models, $\Delta(I_d)$ increases as the sequence length increases. For the shortest sequence length $l=5$, the $\Delta(I_d)\approx 0$, suggesting that at short lengths, (semantic) representational complexity proxies that of a bag of words.}
    \label{fig:emergence_diff_coding}
\end{figure*}

On $d$ and $I_d$ computed for grammars of varying lengths, we found several behaviors that are \emph{emergent} with sequence length:
\begin{enumerate}
    \item Feature complexity increases with sequence length (\Cref{fig:ramping}). This is to be expected, as sequences that are longer contain more information that the LM needs to represent.
    \item Shuffling feature collapse is an emergent property of sequence length. \Cref{fig:emergence_diff_coding} shows the gap between coherent and shuffled $I_d$ grows as sentence length grows. Intuitively, this suggests that phrase-level feature complexity in short coherent sentences proxies that of a bag of words (shuffled). 
\end{enumerate}

\begin{figure*}
    \centering
    \includegraphics[width=0.6\linewidth]{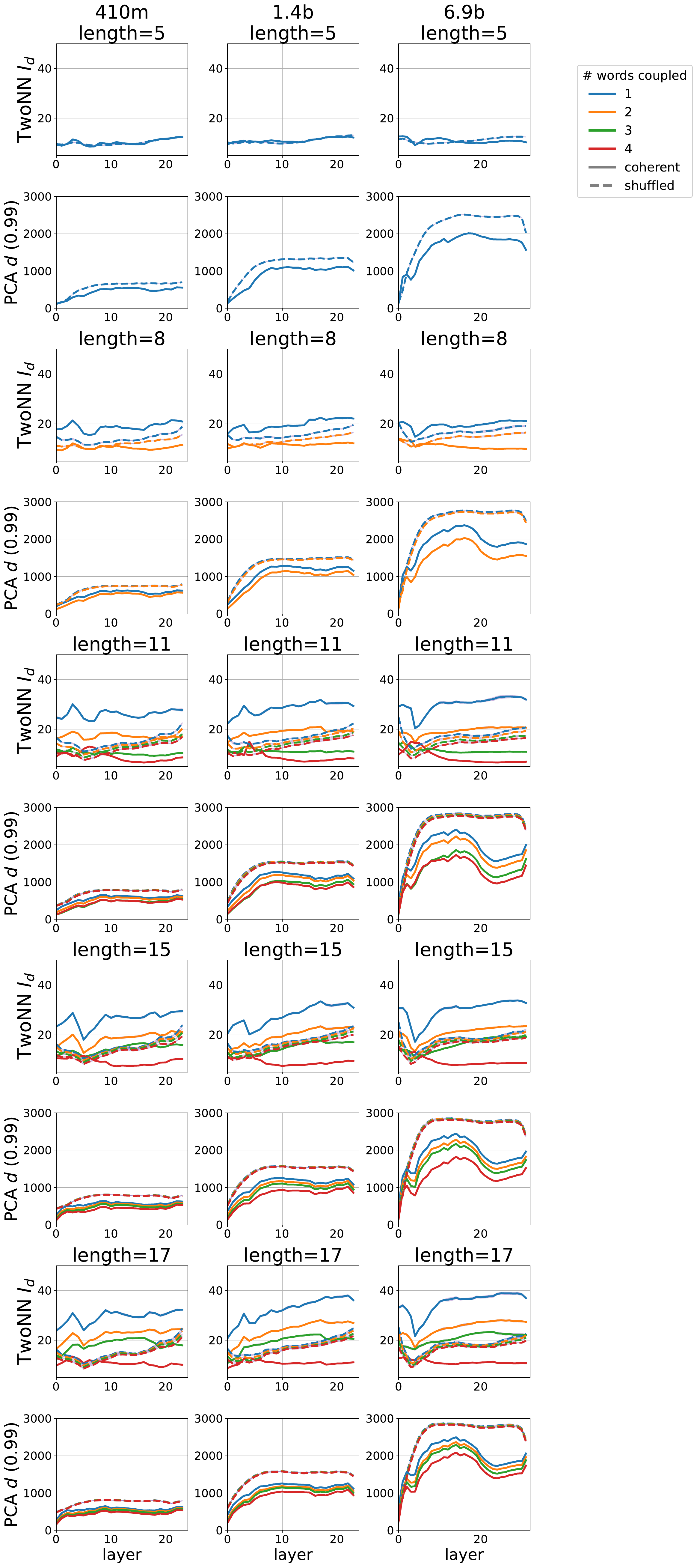}
    \caption{\textbf{Dimensionality over layers, varying sequence length}. Pythia models $\in \{$410m, 1.4b, 6.9b$\}$ (left to right) and sentence lengths $\in \{5, 8, 11, 15, 17\}$, (top to bottom). In all settings, $I_d$ and $d$ monotonically decrease in $k$; upon shuffling, $I_d$ collapses to a low range while $d$ increases.}
    \label{fig:huge_plot}
\end{figure*}

We also find behaviors, corresponding to main article \Cref{sec:id_interpretation}, that are \emph{robust} to sequence length:
\begin{enumerate}
    \item Feature complexity increases as $k$ decreases, no matter the sequence length (\Cref{fig:huge_plot}), showing that results presented in the main generalize to sequences of different length.
    \item Upon shuffling, $I_d$ collapses to a low range, while $d$ increases, no matter the sequence length (\Cref{fig:huge_plot}).
\end{enumerate}

\end{document}